%% file: main.tex
\documentclass[runningheads]{llncs}
\usepackage[mobile]{eccv}

\usepackage{eccvabbrv}
\usepackage{graphicx}
\usepackage{booktabs}
\usepackage[accsupp]{axessibility}
\usepackage[breaklinks,colorlinks,citecolor=eccvblue]{hyperref}
\usepackage{orcidlink}
\usepackage{bbm}
\usepackage[table]{xcolor}
\usepackage{pifont}
\usepackage{multirow}\usepackage{amssymb}
\usepackage{makecell}

\newcommand{\tren}{T-REN}
\newcommand{\boldheader}[1]{\noindent\textbf{#1.}}
\newcommand{\highlight}[1]{\textbf{#1}}
\definecolor{blue}{RGB}{200, 230, 242}
\definecolor{yellow}{RGB}{250, 240, 180}
\newcommand{\cmark}{\checkmark}
\newcommand{\gain}[1]{\textcolor{green!60!black}{$\mathbf{+}$\textbf{#1}}}

\begin{document}
    \title{\tren{}: Learning Text-Aligned Region Tokens Improves Dense Vision-Language Alignment and Scalability} 
    \titlerunning{\tren{}}
    \author{
        Savya Khosla \and
        Sethuraman T V \and
        Aryan Chadha \and
        Alex Schwing \and
        Derek Hoiem
    }
    \authorrunning{S. Khosla et al.}
    \institute{University of Illinois Urbana-Champaign}
    \maketitle
    
    \input{sections/0-abstract}
    \input{sections/1-intro}
    \input{sections/2-related}
    \input{sections/3-method}
    \input{sections/4-experiment}
    \input{sections/5-conclusion}
    
    \bibliographystyle{splncs04}
    \bibliography{main}

    \input{sections/6-supplementary}
\end{document}

%% file: sections/0-abstract.tex
\begin{abstract}

Despite recent progress, vision-language encoders struggle with two core limitations: (1) weak alignment between language and dense vision features, which hurts tasks like open-vocabulary semantic segmentation; and (2) high token counts for fine-grained visual representations, which limits scalability to long videos. This work addresses both limitations. We propose \tren{} (Text-aligned Region Encoder Network), an efficient encoder that maps visual data to a compact set of text-aligned region-level representations (or region tokens).
\tren{} achieves this through a lightweight network added on top of a frozen vision backbone, trained to pool patch-level representations within each semantic region into region tokens and align them with region-level text annotations.
With only 3.7\% additional parameters compared to the vision-language backbone, this design yields substantially stronger dense cross-modal understanding while reducing the token count by orders of magnitude. Specifically, \tren{} delivers +5.9 mIoU on ADE20K open-vocabulary segmentation, +18.4\% recall on COCO object-level text-image retrieval, +15.6\% recall on Ego4D video object localization, and +17.6\% mIoU on VSPW video scene parsing, all while reducing token counts by more than 24$\times$ for images and 187$\times$ for videos compared to the patch-based vision-language backbone. The code and model are available at \url{https://github.com/savya08/T-REN}.

\keywords{Vision-language encoders \and Region encoder network \and Scalable representations}

\end{abstract}

%% file: sections/1-intro.tex
\section{Introduction}
\label{sec:intro}

\begin{figure}[t]
    \centering
    \includegraphics[width=\linewidth]{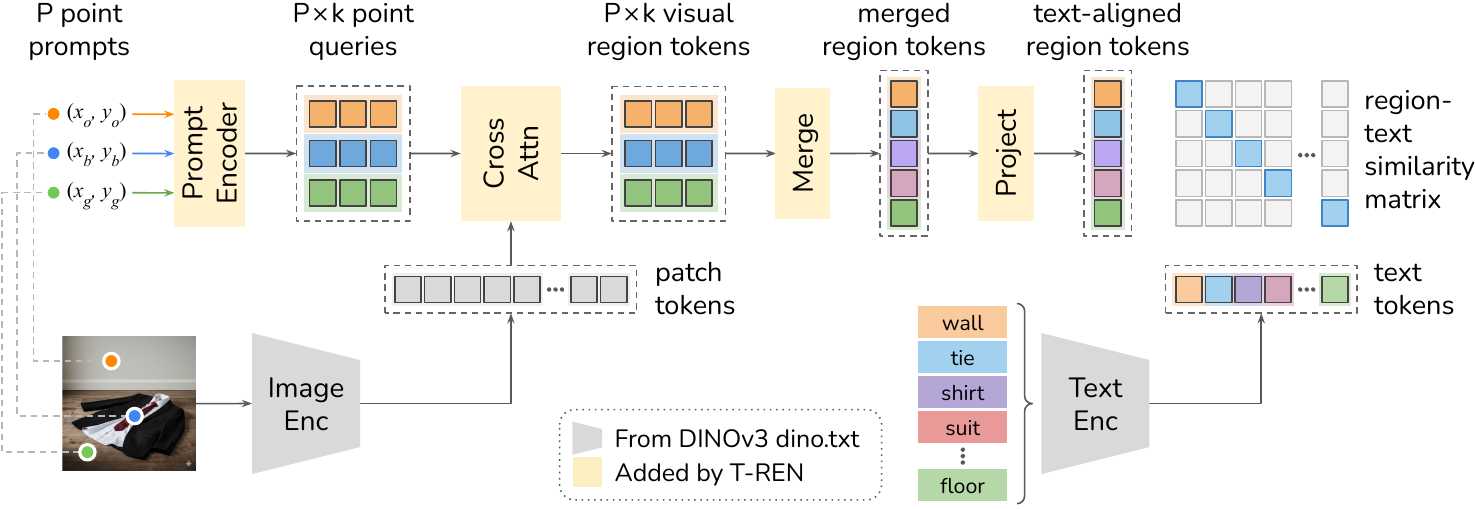}
    \vspace{-1.8em}
    \caption{{\bf Overview of \tren{}.} Given an input image and $P$ point prompts, \tren{} pools semantically related patch features via cross-attention to produce $k=3$ region tokens per point prompt. Since multiple tokens may capture overlapping semantics (e.g., when points lie on the same region), highly similar tokens are merged to reduce redundancy. The resulting region tokens are projected into the text embedding space and matched with text encodings using cosine similarity for open-vocabulary tasks.}
    \label{fig:overview}
    \vspace{-1.5em}
\end{figure}

Despite strong performance on global image-text tasks, modern vision-language encoders~\cite{CLIP,SigLIP,SigLIP2,PerceptionEncoder,DINOtxt} remain bottlenecked by two structural limitations. First, cross-modal alignment between language and dense visual features remains weak, hindering open-vocabulary semantic segmentation, retrieval, and localization. Second, patch-based visual representations generate thousands of tokens per image, leading to substantial memory and compute overhead as visual input length increases. Together, these issues limit the fine-grained understanding and scalability of current vision-language systems.

Existing approaches address these challenges in isolation. Token compression methods reduce the number of visual tokens but often sacrifice downstream performance~\cite{ToMe,DynamicViT,AViT,EViT}. Conversely, methods that improve dense alignment continue to operate on patch-level representations~\cite{MaskCLIP,SCLIP,ClearCLIP,ProxyCLIP}, inheriting the inefficiencies and semantic fragmentation of patch tokens. We argue that patch-level tokens are a suboptimal representational unit for dense vision-language modeling: they are too fine to be semantically meaningful, yet too numerous to scale efficiently.

We address both challenges simultaneously by shifting the unit of representation from patches to regions. We introduce \tren{}, a Text-aligned Region Encoder Network that converts patch features from a vision backbone into a compact set of region tokens and aligns them with region-level text annotations. Jointly learning region-based pooling and region-text alignment produces stronger dense cross-modal representations while dramatically reducing the token count. We use DINOv3-based dino.txt~\cite{DINOtxt} as the pretrained backbone. 

Inspired by REN~\cite{REN}, \tren{} uses a cross-attention module in which a grid of point prompts serves as queries over patch features from a frozen vision backbone. The attention operation pools patch features that correspond to the same semantic unit into region tokens. These region tokens are then projected into the text embedding space for alignment with region-level text annotations. \Cref{fig:overview} provides an overview. Building on REN, we introduce a key refinement: instead of generating a single region token per point prompt, \tren{} produces multiple tokens per prompt. This richer encoding captures both whole objects and their parts, alleviating the part-whole ambiguity inherent in REN and improving retrieval.

We evaluate \tren{} on open-vocabulary semantic segmentation and retrieval over both images and videos. Our results show that \tren{} consistently improves dense task performance while substantially reducing the number of visual tokens. Importantly, these gains come at minimal cost, requiring only 3.7\% additional parameters on top of the original patch-based vision-language backbone.

In summary, our contributions are:
\begin{enumerate}
    \item {\bf Improving cross-modal alignment.} By jointly learning the spatial pooling of patch features into region tokens and region-level text alignment, \tren{} enhances dense vision-language understanding. Pooling and alignment associate text annotations with a precise visual region, enabling more accurate open-vocabulary segmentation (+5.9 mIoU on ADE20K, +15.8 on Cityscapes, and +17.6\% on VSPW) and retrieval (+18.4\% recall on COCO and +15.6\% recall on Ego4D).
    
    \item {\bf Making visual encoding more scalable.} Adding only a small network on top of the pretrained backbone (+3.7\% parameters), \tren{} dramatically reduces visual token counts (e.g., 24.4$\times$ for COCO images, 187.5$\times$ for Ego4D videos, and 254.5$\times$ for VSPW videos). This enables the processing of long videos and large image collections, which would otherwise be computationally prohibitive.

    \item {\bf Enabling more expressive region encoding.} \tren{} addresses the part-whole ambiguity present in its predecessor, REN. While REN generates a single region token per point prompt, \tren{} produces multiple region tokens for each prompt, capturing both fine-grained object parts and the full object instance. This yields richer representations and improves retrieval (+5.7\% recall on COCO).
\end{enumerate}

%% file: sections/2-related.tex
\section{Related Work}
\label{sec:related}

\boldheader{Vision-language encoders}
The seminal CLIP~\cite{CLIP} popularized large-scale contrastive image-text pretraining for open-vocabulary vision understanding. Several subsequent works have proposed improvements to this paradigm. For example, SigLIP~\cite{SigLIP} and SigLIP-2~\cite{SigLIP2} replace the contrastive objective with a sigmoid loss to improve pretraining efficiency, and PE~\cite{PerceptionEncoder} performs large-scale contrastive pretraining with joint image-video training to improve cross-modal alignment. In parallel, dino.txt~\cite{DINOtxt} adapts powerful self-supervised vision backbones such as DINOv2~\cite{DINOv2} and DINOv3~\cite{DINOv3} for vision-language tasks, improving dense open-vocabulary understanding. \tren{} builds on DINOv3-based dino.txt by introducing a lightweight region pooling and text-alignment module without retraining the underlying encoder.

\vspace{0.6em}
\boldheader{Improving dense alignment}
CLIP-based models are optimized for image-level understanding, leaving patch-level representations weakly aligned with language. To address this, training-free methods extract denser signals by modifying CLIP's attention at inference~\cite{MaskCLIP,SCLIP,ClearCLIP} or propagate spatial priors from strong vision models (e.g., DINO~\cite{DINO} or SAM~\cite{SAM}) into CLIP's feature space~\cite{CLIP-DINOiser,ProxyCLIP}. In parallel, supervised methods enhance CLIP with region-level annotations through masked-region fine-tuning~\cite{OVSeg}, region-text contrastive pretraining~\cite{CLOC}, or bounding-box alignment with detailed captions and hard negatives~\cite{FG-CLIP}. These methods, however, operate on patch-level tokens, distributing object information across many tokens without semantic grouping. In contrast, \tren{} pools patches into region tokens and jointly aligns them with text, producing semantically grounded representations.

\vspace{0.6em}
\boldheader{Reducing the visual token count}
Patch-based vision encoders produce hundreds or thousands of tokens per image, regardless of visual content, leading to substantial memory and compute overhead at scale. One line of work addresses this by pruning low-salience tokens at inference using predicted importance scores, attention weights, or diversity criteria~\cite{DynamicViT,AViT,FastV,PyramidDrop}. A complementary approach aggregates patch tokens using similarity-based pooling or cross-attention with a small set of learned query tokens~\cite{ToMe,EViT,Perceiver,Q-Former}. For video, temporally redundant tokens across frames are seldom merged to keep sequence length tractable~\cite{DyCoke,LongVU}. These methods generally trade a performance drop for efficiency. T-REN addresses this trade-off by producing tokens that are compact by construction: each region token pools an entire semantic unit, naturally yielding far fewer tokens without discarding spatial information.

\vspace{0.6em}
\boldheader{Region-based representations}
Shlapentokh-Rothman et al.~\cite{RegionBasedRep} show that combining SAM~\cite{SAM} masks with DINOv2~\cite{DINOv2} features produces compact and effective region representations for segmentation and retrieval, though the SAM segmentation step adds substantial overhead. REN~\cite{REN} addresses SAM's cost with a lightweight cross-attention module that generates region tokens from point-prompt queries over frozen patch features, running $60\times$ faster than SAM-based pipelines. However, REN generates only a single token per prompt, leading to part-whole ambiguity, and its tokens are not trained for text alignment. \tren{} addresses these limitations by producing multiple tokens per prompt to capture hierarchical structure and jointly learning region pooling with text alignment using a DINOv3-based dino.txt backbone.

%% file: sections/3-method.tex
\section{\tren{}}
\label{sec:method}

\begin{figure}[t]
    \centering
    \includegraphics[width=0.75\linewidth]{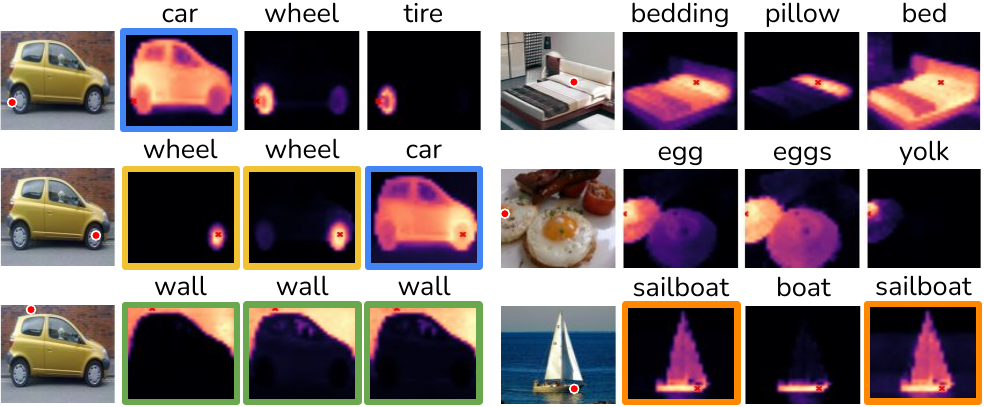}
    \vspace{-0.7em}
    \caption{{\bf Qualitative examples of \tren{}'s cross-attention masks.} Each point prompt (shown in red) produces $k=3$ cross-attention masks that pool patch tokens into region tokens. Tokens corresponding to masks covering the same semantic region are subsequently merged into a single region token. In the yellow-car example (left), masks highlighted with a blue border merge into a single \emph{car} token, yellow masks into a \emph{wheel} token, and green masks into a \emph{wall} token.}
    \label{fig:cross-attention-masks}
    \vspace{-1.5em}
\end{figure}

\Cref{fig:overview} provides an overview of \tren{}. Our objective is two-fold: (1) improve fine-grained alignment between vision and language, and (2) reduce the token budget required to represent visual content. To achieve this, \tren{} pools dense patch features from the DINOv3 ViT-L image encoder~\cite{DINOv3} into a compact set of semantically meaningful region tokens and aligns these tokens with text embeddings produced by the DINOv3-based dino.txt text encoder~\cite{DINOtxt}. The pipeline and model architecture are described in \Cref{sec:approach}, and the training objectives are detailed in \Cref{sec:training}.

\subsection{Generating Compact Set of Text-Aligned Region Tokens}
\label{sec:approach}

\boldheader{Encode point prompts into point queries}
Following REN~\cite{REN}, \tren{} employs point prompts to query and pool patch tokens into region tokens. Because a single spatial location may correspond to multiple semantic entities (e.g., overlapping objects or part–whole structures; see the blue prompt in \Cref{fig:overview}), we generate $k$ region tokens per prompt. Our prompt encoder transforms each prompt $(x_p, y_p)$ into $k=3$ {\em point queries} using learned token embeddings. Specifically, the 2D position of the prompt is encoded with Gaussian Random Fourier Feature (RFF) embeddings and added to the learned tokens, yielding $k$ distinct queries per location. To ensure full coverage of the image, we use a 2D grid of $P$ prompts, resulting in $P \times k$ point queries in total.

\vspace{0.6em}
\boldheader{Pool patch tokens into visual region tokens using point queries}
The point queries are processed by a stack of $L=2$ decoder layers. Each layer consists of two stages: (1) cross-attention between the point queries and the patch tokens, enabling each point query to gather spatially relevant visual information; and (2) self-attention among the $k$ point queries associated with the same spatial location, allowing interactions among  candidate entities that arise from a single prompt. To preserve spatial grounding across decoder layers, we re-inject the 2D positional encoding of each prompt after every self-attention block before passing the queries to the next layer. After $L$ such layers, the processed queries are converted into {\em visual region tokens} using a final cross-attention operation over the patch tokens. This final cross-attention layer uses a single attention head and omits value and output projections, ensuring that the pooled features remain in the feature space of the frozen vision backbone. Due to the single-head design and our learning objective of pooling patch tokens within each region, the attention weight $a_i = \text{softmax}(q_i K^T / \sqrt{d_k})$ of a point query $q_i$ naturally resembles a low-resolution region mask. Thus, each of the $P \times k$ point queries produces a {\em cross-attention mask} (\Cref{fig:cross-attention-masks}) that pools patch tokens into a corresponding region token, yielding $P \times k$ visual region tokens in total.

\vspace{0.6em}
\boldheader{Merge similar visual region tokens}
While a dense prompt grid ensures spatial coverage, it also introduces redundancy. For example, in \cref{fig:overview}, all point prompts placed on the wall will produce similar region tokens, as they correspond to the same semantic entity without any part-whole structure. To mitigate this redundancy, we merge highly similar visual region tokens. Specifically, we average-pool tokens whose pairwise cosine similarity exceeds $\tau_\text{token}=0.975$ or whose cross-attention mask IoU exceeds $\tau_\text{mask}=0.8$. This simple merging strategy works because our training encourages tokens from the same semantic region to be identical (see \Cref{sec:training} for more details). After merging, we obtain $M$ {\em merged region tokens} representing the image, where $M$ adapts to the visual complexity: visually sparse scenes yield fewer tokens, while cluttered scenes produce more tokens. We also cache the point-prompt coordinates of the merged region token, enabling association between region tokens and their spatial locations.

\vspace{0.6em}
\boldheader{Align merged region tokens with text}
Finally, we project the merged region tokens into the text embedding space using an MLP. The resulting text-aligned region tokens can be matched with text encodings via cosine similarity for open-vocabulary tasks.

\vspace{0.6em}
\boldheader{Temporal aggregation for videos}
For long videos, even a compressed set of per-frame region tokens results in an excessive number of tokens, especially for applications like episodic memory storage or streaming. We therefore extend our merging strategy temporally by aggregating region tokens across consecutive frames into {\em track tokens}. At each frame $t+1$, we compare its merged region tokens (prior to text projection) to the active track tokens accumulated up to frame $t$, where a track is considered active at time $t$ if its object appears in frame $t$. We compute pairwise cosine similarity between the frame tokens and active track tokens and retain all pairs whose similarity exceeds $\tau_\text{track}=0.65$. Then, greedy one-to-one matching is performed: each token from frame $t+1$ is assigned to the most similar active track. Tokens that do not match any active track initialize new tracks. This process is repeated sequentially across all frames in a streaming fashion, producing a set of object tracks. For each track, the constituent tokens are average pooled to obtain a single track token. We also cache the associated frame indices for each track token, preserving temporal grounding.

\subsection{Training via Contrastive and Distillation Objectives}
\label{sec:training}

Our training objectives are designed around two complementary goals: (1) learning region tokens that capture object- and part-level semantics, and (2) preserving the rich representation space of the frozen vision-language backbone. To achieve the first goal, we apply contrastive losses in both the visual and text-aligned feature spaces. In the visual space, the contrastive objective (\cref{eq:vis-cont-loss}) encourages region tokens originating from the same ground-truth mask to cluster together while pushing apart tokens from different regions, directly facilitating the similarity-based grouping that underlies our token merging strategy. In the text-aligned space, the contrastive objective (\cref{eq:txt-cont-loss}) aligns region tokens with their corresponding category-level text encodings, improving region-level open-vocabulary recognition. To achieve the second goal, we apply distillation losses (\cref{eq:txt-dist-loss}) that anchor both the visual and text-aligned region tokens to their respective targets obtained by mask-pooling features from the frozen backbone. Visual distillation prevents the learned tokens from drifting away from the pretrained feature space, while text-aligned distillation preserves the open-vocabulary capability inherited from the vision-language backbone. Together, these four objectives ensure that region tokens are semantically coherent, groupable by similarity, and remain grounded in the backbone's representation space. We additionally supervise the cross-attention masks to match ground-truth masks (\cref{eq:attn-mask-loss}), which we find accelerates convergence without affecting the final performance. Training uses a mixture of five segmentation datasets with biased point sampling to increase supervision from spatially overlapping regions, and Hungarian matching to handle the variable number of regions per prompt.

\vspace{0.6em}
\boldheader{Data and Supervision Signal}
\tren{} is trained on a mixture of five segmentation datasets: COCOStuff~\cite{COCOStuff}, OpenImagesV7~\cite{OpenImagesv7}, PhraseCut~\cite{PhraseCut}, Mapillary~\cite{Mapillary}, and SA-1B~\cite{SAM}. For each training image, point prompts are sampled from locations inside ground-truth segmentation masks. To emphasize points that overlap multiple semantic entities (e.g., part-whole regions), sampling probability is proportional to the square of the number of overlapping regions. Distillation targets for visual region tokens are obtained by average pooling DINOv3 features within each corresponding mask. Targets for the text-aligned region tokens are obtained by processing the visual targets using the text-alignment vision block from DINOv3-based dino.txt~\cite{DINOtxt,DINOv3}.

\vspace{0.6em}
\boldheader{Hungarian matching}
Each point prompt may have up to $k=3$ target regions, while \tren{}'s cross-attention module always predicts $k$ visual region tokens per prompt. To align predictions with target regions, we construct a cost matrix using cosine distances between predicted and target visual region tokens and solve a one-to-one assignment with the Hungarian algorithm~\cite{DETR}. Unmatched predicted tokens (when fewer than $k$ targets exist) are excluded from the loss. This ensures permutation-invariant training and flexible assignment of predicted tokens to semantic entities.

\vspace{0.6em}
\boldheader{Contrastive token learning}
Using the Hungarian assignment, matched region tokens are supervised with contrastive objectives in both visual and text-aligned spaces. Formally, let $r_i^{(v)}$ and $r_i^{(t)}$ denote a normalized visual and text-aligned region tokens, $m_i$ the corresponding region mask, and $t_i$ the text encoding of the corresponding annotation. The contrastive losses for a batch of $N$ regions are then computed as:
\vspace{-1em}

\begin{equation}
    \vspace{-1em}
    \mathcal{L}_{\text{cont}}^{(v)} = -\frac{1}{N} \sum_{i=1}^{N} \log \frac{\sum\limits_{j=1}^{N} \mathbbm{1}_{[j \ne i,\  m_j = m_i]} e^{r_i^{(v)} \cdot r_j^{(v)} / \tau}}{\sum\limits_{k=1}^{N} \mathbbm{1}_{[k \ne i]} e^{r_i^{(v)} \cdot r_k^{(v)} / \tau}},
    \label{eq:vis-cont-loss}
\end{equation}
\begin{equation}
    \mathcal{L}_{\text{cont}}^{(t)} = -\frac{1}{2N} \sum_{i=1}^{N} \Bigg( \log \frac{e^{r_i^{(t)} \cdot t_i / \tau}}{\sum\limits_{k=1}^{N} \mathbbm{1}_{[t_k \ne t_i]} e^{r_i^{(t)} \cdot t_k / \tau}} + \log \frac{e^{r_i^{(t)} \cdot t_i / \tau}}{\sum\limits_{k=1}^{N} \mathbbm{1}_{[t_k \ne t_i]} e^{r_k^{(t)} \cdot t_i / \tau}} \Bigg).
    \label{eq:txt-cont-loss}
\end{equation}

\vspace{0.6em}
\boldheader{Distillation loss}
To keep region tokens aligned with the pretrained backbone, we apply a cosine-based distillation loss in both visual and text-aligned spaces, encouraging predicted tokens to remain close to their Hungarian-assigned targets. Specifically, if $\Tilde{r}_i^{(v)}$ and $\Tilde{r}_i^{(t)}$ denote the visual and text-aligned targets,

\begin{equation}
    \mathcal{L}_{\text{dist}} = 
    \frac{1}{N} \sum_{i=1}^{N} 
    \bigg[ 
    \bigg( 1 - \frac{r_i^{(v)} \cdot \Tilde{r}_i^{(v)}}{\|r_i^{(v)}\| \|\Tilde{r}_i^{(v)}\|} \bigg) 
    +
    \bigg( 1 - \frac{r_i^{(t)} \cdot \Tilde{r}_i^{(t)}}{\|r_i^{(t)}\| \|\Tilde{r}_i^{(t)}\|} \bigg) 
    \Bigg].
    \label{eq:txt-dist-loss}
\end{equation}

\vspace{0.6em}
\boldheader{Attention supervision}
Finally, to accelerate training, each max-normalized cross-attention mask $\Tilde{a}_i$ is supervised to match the ground-truth masks $m_i$ using a combination of binary cross-entropy loss $\ell_{\text{bce}}$ and DICE loss $\ell_{\text{dice}}$:
\begin{equation}
    \mathcal{L}_{\text{attn}} = \frac{1}{N} \sum_{i=1}^{N} \big[ \ell_{\text{bce}}(\Tilde{a}_i, m_i) + \ell_{\text{dice}}(\Tilde{a}_i, m_i) \big].
    \label{eq:attn-mask-loss}
\end{equation}

%% file: sections/4-experiment.tex
\section{Experiments}
\label{sec:experiments}

We evaluate \tren{} across three settings: zero-shot object-level retrieval over image databases (\Cref{sec:retrieval}), open-vocabulary semantic segmentation (\Cref{sec:segmentation}), and object localization and scene parsing in videos (\Cref{sec:videos}). We do not evaluate on global image-level tasks (e.g., image classification or caption-based retrieval), as \tren{} leaves the backbone’s image-level representation unchanged. Consequently, for such tasks, \tren{} preserves the original performance of the underlying DINOv3-based dino.txt model. Additional ablations and analyses are presented in \Cref{sec:ablations}. All experiments are conducted in a zero-shot manner. 
\vspace{-0.5em}
\subsection{Retrieval}
\label{sec:retrieval}

\input{tables/visual-haystacks}

We evaluate \tren{} on the Visual Haystacks Single-Needle Challenge, where the task is: given a database of $D$ images, answer queries of the form {\em ``For the image containing the [anchor object], is there a [target object]?''}.

\tren{} addresses this task using a simple two-step procedure. First, we retrieve the image containing the anchor object by computing the cosine similarity between the anchor text embedding and all text-aligned region tokens across the $D$ images and selecting the image whose region token achieves the highest similarity score. Second, we determine whether the retrieved image contains the target object by computing the cosine similarity between the target text embedding and all region tokens within that image. We answer yes if the maximum similarity exceeds a threshold of $\tau_\text{sim} = 0.23$, and no otherwise. 

The results are summarized in \Cref{tab:haystacks}. \tren{} consistently outperforms vision-language encoders (including patch-based models such as DINOv3 dino.txt and region-based models such as REN), as well as open-source MLLMs and retrieval-augmented methods across all values of $D$. It further surpasses Gemini-1.5 Pro and GPT-4o across all evaluated scales, while showing competitive performance with Gemini-3 Pro. Importantly, \tren{} achieves these gains while incurring substantially lower computational costs than MLLMs.

In \Cref{fig:text-based-retrieval}, we analyze \tren{}'s performance on text-based retrieval (which is the first step in the Visual Haystacks approach). Compared to its vision-language backbone, \tren{} improves recall (R@1) by an average of 18.4\% across values of $D$. The performance gap widens as $D$ increases. For example, at $D=500$, the improvement reaches 28.6\%. Notably, this gain is achieved while using $24.4\times$ fewer tokens to represent the database on average across values of $D$.

\input{tables/retrieval-and-segmentation}
\vspace{-0.5em}

\subsection{Open-Vocabulary Semantic Segmentation}
\label{sec:segmentation}
We evaluate \tren{} on Open-Vocabulary Semantic Segmentation (OVSS), where the goal is to assign a semantic label to each pixel of an image in a zero-shot manner. 

We prompt \tren{} with a $24 \times 24$ grid of points and obtain $k=3$ text-aligned region tokens per point. We then average-pool the $k$ tokens at each point and compute cosine similarity against text encodings for all $C$ classes in the dataset. This produces a $24 \times 24 \times C$ logit map, which is upsampled to the original image resolution of $384 \times 384 \times C$, and the most similar class is taken as the prediction for each pixel. To better assess fine-grained alignment at the point level, we disable token merging for this evaluation. This evaluation protocol follows the standard OVSS setup used to assess vision-language encoders~\cite{DINOv3}.

The results are reported in \Cref{tab:semantic-segmentation}. Compared to the DINOv3-based dino.txt, \tren{} improves performance by +5.9 mIoU on ADE20k~\cite{ADE20k} and +15.8 mIoU on Cityscapes~\cite{Cityscapes}, demonstrating substantially stronger dense vision–language alignment. As shown in \Cref{fig:segmentation-examples}, \tren{}’s zero-shot segmentations adhere more closely to object boundaries, explaining the observed improvement. \tren{} also surpasses recent SAM-guided OVSS approaches~\cite{TextRegion,Trident,RADSeg}, which rely on an additional segmentation model (SAM~\cite{SAM}) for mask refinement and operate at higher input resolutions. Without any external refinement, \tren{} already achieves superior performance at 384p. Furthermore, its accuracy improves consistently as input resolution increases (see \Cref{fig:segmentation-resolution}).

\vspace{-0.5em}
\subsection{Scaling to Video}
\label{sec:videos}

Existing approaches to long-video tasks typically rely on either representing each frame with a single global token or maintaining dense patch-level representations while aggressively subsampling frames to control sequence length. Both strategies impose inherent trade-offs. Global frame tokens lack the spatial granularity needed to capture small objects in cluttered scenes, while temporal subsampling risks missing frames in which short-lived objects appear. In this section, we show that track tokens from \tren{} provide an effective alternative: they preserve fine-grained spatial information by focusing on semantically meaningful regions while maintaining a compact token budget suitable for long video sequences. Consequently, \tren{} yields consistent improvements in both performance and efficiency for retrieval and segmentation in the video setting.

\vspace{0.6em}
\boldheader{Query localization in long videos}
We evaluate \tren{} on the task of localizing the last occurrence of an object in long episodic memory videos from Ego4D. Given a video and a query object, the goal is to identify the temporal window corresponding to the object's final appearance. The query is provided both as a text prompt and as a visual crop of the object. The videos average 140 seconds in duration and are sampled at 5 FPS. The target temporal window for the object's final occurrence spans 3 seconds on average.

To efficiently localize the queried object, we match video track tokens to the query by combining visual and textual similarity. Formally, we compute:
\vspace{-0.5em}
\[
\texttt{track-similarity} = \texttt{visual-similarity} \times \texttt{textual-similarity}, 
\vspace{-0.55em}
\]
where \texttt{visual-similarity} is the cosine similarity between the visual query tokens and the video's visual track tokens (i.e., track tokens obtained by temporally aggregating per-frame visual region tokens), and \texttt{textual-similarity} is the cosine similarity between the query text encoding and the video's text-aligned track tokens. We then retain all tracks whose similarity exceeds $\tau_\text{sim} = 0.18$ and report the temporal window of the last-ending track as the final prediction.

\Cref{tab:video-tasks} summarizes the results. Compared to DINOv3-based dino.txt, \tren{} improves query recall by 15.6\% while using $187.5\times$ fewer tokens to represent a video on average. This substantial reduction in token count yields important practical benefits. For example, in our evaluation with Ego4D, patch-based representations of long videos exceeded the memory capacity of a single NVIDIA A40 GPU and required streaming-based processing. On the other hand, \tren{}, owing to its $187\times$ compression, faces no such bottleneck: representations of even an 8-minute video (2400 frames) fit entirely within the memory of a single NVIDIA A40. These results highlight that \tren{} is particularly well suited for episodic memory retrieval, where video representations must be stored efficiently on disk, fit within limited GPU memory, or be deployed on edge devices.

\input{tables/video-tasks}

\vspace{0.6em}
\boldheader{Video scene parsing}
We evaluate \tren{} on video scene parsing, where the goal is to assign a semantic label to every pixel in every frame of a video sequence.

To efficiently control the token budget for videos, we leverage temporally aggregated track tokens. Specifically, we compute the cosine similarity between each text-aligned track token and the text embeddings of all category labels, and assign the category with the highest similarity to the track. The predicted label for a track token is then applied to all spatio-temporal regions contributing to that track. Concretely, each track token is formed by first spatially merging point-prompted region tokens within a frame and then temporally aggregating similar tokens across frames; the assigned category is propagated to all spatial locations and time steps associated with that track.

\Cref{tab:video-tasks} summarizes the results. \tren{} surpasses both DINOv3-based dino.txt and REN while using $254.5\times$ and $11.1\times$ fewer tokens per video, respectively. We also analyze the effect of merging region tokens within and across frames (\Cref{tab:merging-effect}) and find that our merging strategy significantly reduces the token count without degrading representation quality (see \Cref{sec:ablations} for more details).

\vspace{-1em}
\subsection{Ablations}
\label{sec:ablations}
\begin{figure}[tb]
    \centering
    \begin{subfigure}{0.32\linewidth}
        \includegraphics[width=\linewidth]{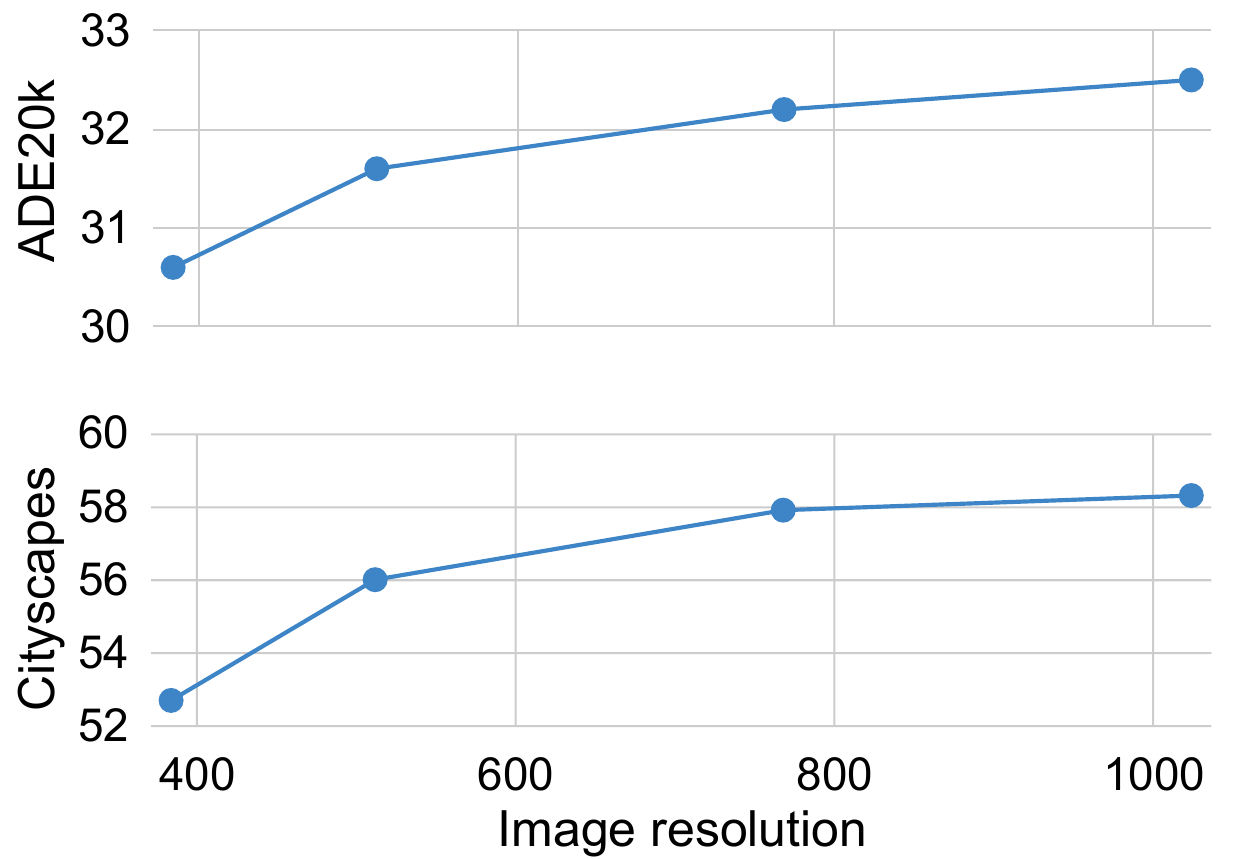}
        \caption{}
        \label{fig:segmentation-resolution}
    \end{subfigure}
    \hfill
    \begin{subfigure}{0.32\linewidth}
        \includegraphics[width=\linewidth]{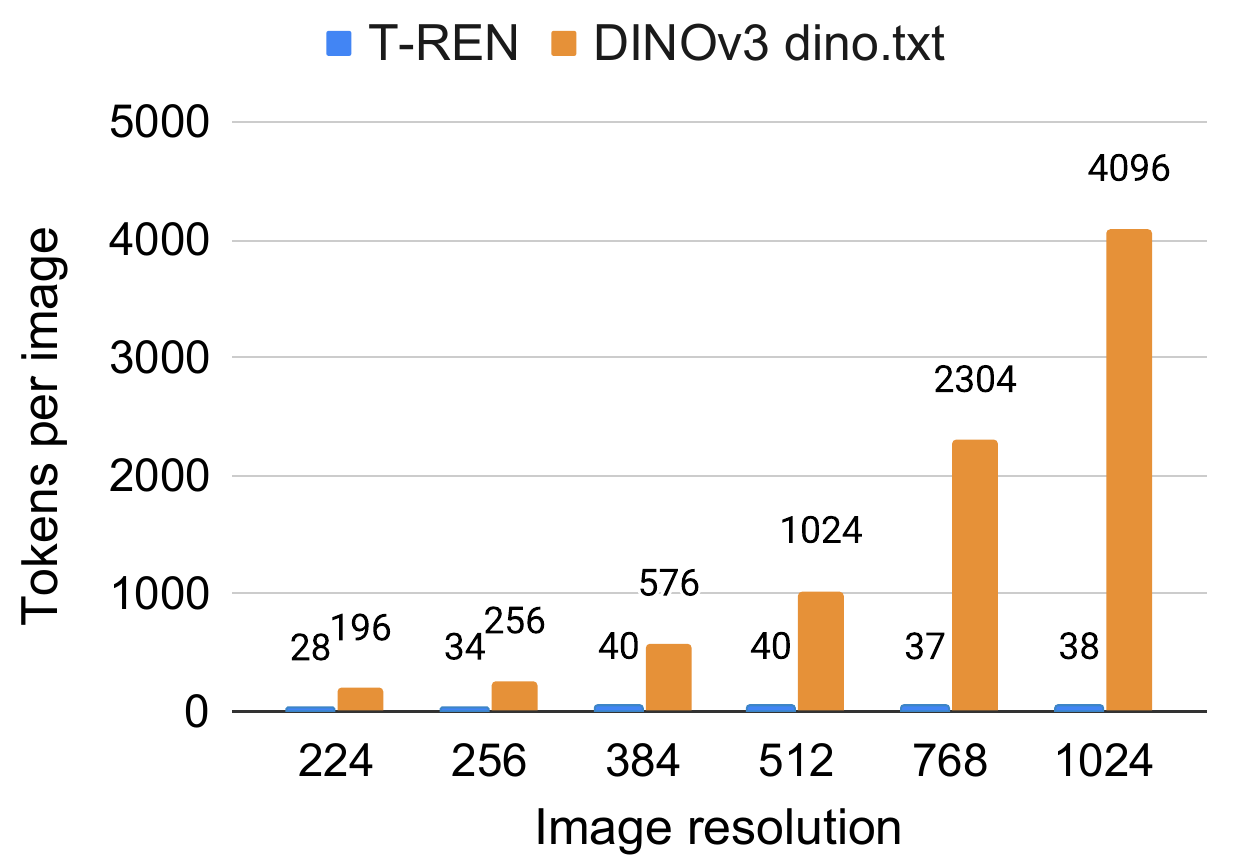}
        \caption{}
        \label{fig:token-resolution}
    \end{subfigure}
    \hfill
    \begin{subfigure}{0.32\linewidth}
        \includegraphics[width=\linewidth]{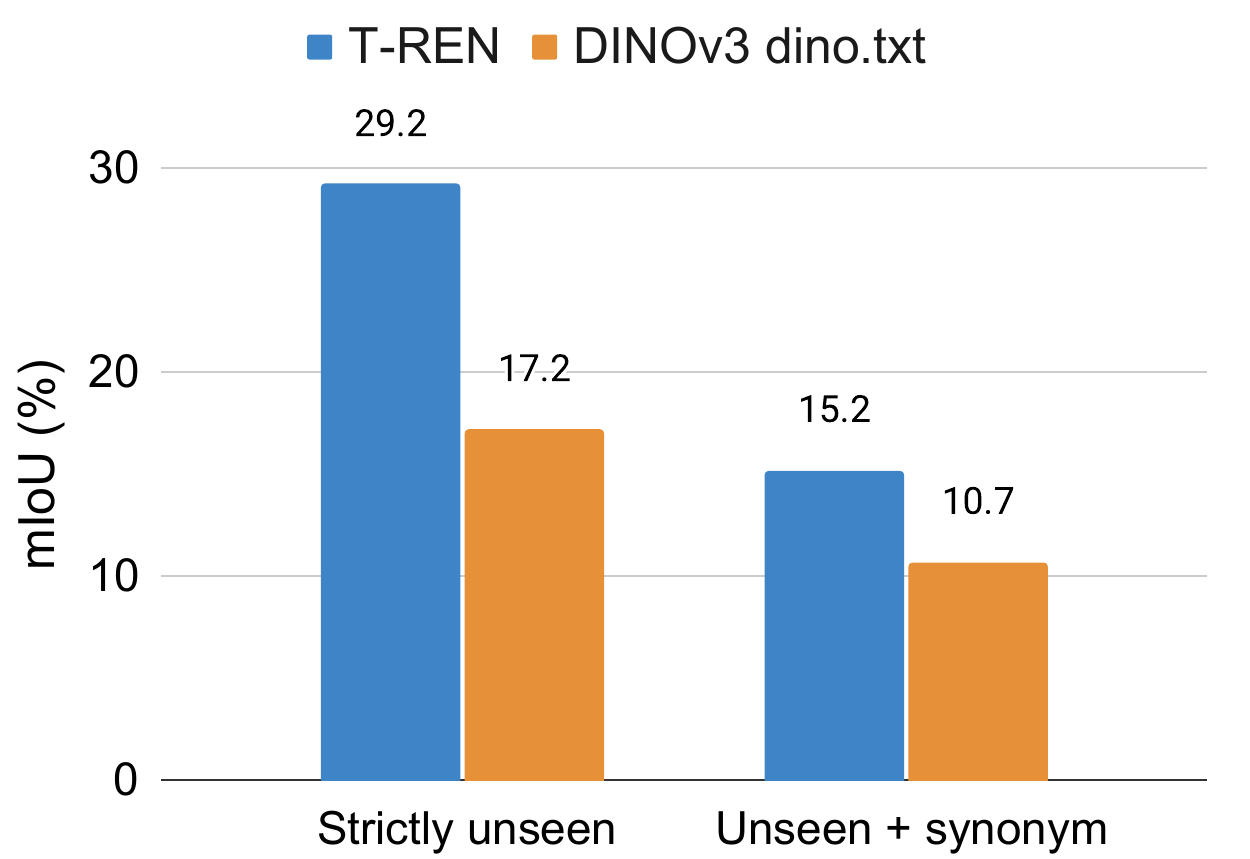}
        \caption{}
        \label{fig:generalization}
    \end{subfigure}
    \vspace{-0.7em}
    \caption{{\bf Resolution scaling and generalization.} (a) Segmentation mIoU improves as input resolution increases. (b) Unlike patch-based encoders, token count of \tren{} remains nearly constant as the image resolution is increased. (c) \tren{} generalizes effectively to text classes unseen during training.}
    \label{fig:analysis}
    \vspace{-1.5em}
\end{figure}

We perform ablations to validate our core design choices. First, we show that jointly learning spatial pooling and text alignment is critical for strong fine-grained vision–language alignment (\Cref{tab:ablation}). Next, we demonstrate that merging region tokens within and across frames removes redundancy in video representations without degrading quality (\Cref{tab:merging-effect}). We then highlight that multi-region token prediction is essential for learning expressive, hierarchically consistent region-based representations (\Cref{fig:multi-token}). In each of these studies, we modify only a single component while keeping the rest of the architecture and training protocol fixed. Finally, we analyze the impact of input resolution on \tren{} and its generalization to classes unseen during training (\Cref{fig:analysis}).

\vspace{0.6em}
\boldheader{Ablating region pooling}
We isolate the effect of region pooling by training a variant that bypasses spatial pooling and directly aligns patch-level features with region-level text annotations. Specifically, each patch token from the DINOv3 backbone is projected into the text embedding space and supervised using the annotation of the region in which it resides. As shown in \Cref{tab:ablation}, this variant underperforms \tren{}, demonstrating that pooling patch tokens into region tokens not only leads to a reduced token count but also improves dense vision-language alignment.

\input{tables/ablation}

\vspace{0.6em}
\boldheader{Ablating text-alignment}
We next evaluate the importance of jointly learning text alignment with region pooling. To this end, we train a variant that learns only region pooling and derives text-aligned region tokens post hoc. Specifically, we use the learned cross-attention masks to pool text-aligned patch features from the vision side of DINOv3-based dino.txt into region tokens. As shown in \Cref{tab:ablation}, this decoupled strategy yields substantially degraded performance. We attribute this to the spatial noise in independently learned text-aligned patch features, which often fail to respect object boundaries (see \Cref{fig:segmentation-examples}). Consequently, although the region pooling module learns precise region assignments, applying them to spatially imprecise patch features produces misaligned semantic representations.

\vspace{0.6em}
\boldheader{Ablating token merging}
We analyze the impact of the proposed token merging stages, which consist of: (1) merging similar tokens produced by different point queries within a frame and (2) merging similar region tokens across frames. We measure their effect on the VSPW video scene parsing task, with results shown in \Cref{tab:merging-effect}. Both merging steps preserve task performance with negligible degradation while drastically reducing the number of tokens, indicating that the removed redundancy carries negligible discriminative information for this task.

\vspace{0.6em}
\boldheader{Ablating multi-region token prediction per point prompt}
A fundamental architectural difference between REN~\cite{REN} and \tren{} is that \tren{} predicts multiple region tokens for each point prompt. To assess the impact of this design upgrade, we train a variant of \tren{} that predicts only a single token per point prompt, keeping all other components unchanged. As shown in \Cref{fig:multi-token}, the single-token variant consistently underperforms the proposed multi-token setup in zero-shot retrieval and classification. This degradation indicates that constraining each point to a single token limits the model’s ability to represent multiple valid hierarchical interpretations associated with a location (e.g., an object part and the full instance). Allowing multiple tokens per point preserves this part-whole structure and leads to more expressive region-level representations.

\input{tables/merging-effect}

\vspace{0.6em}
\boldheader{Effect of image resolution}
Performance on vision tasks generally improves with higher input resolution, as illustrated for OVSS in \Cref{fig:segmentation-resolution}. For patch-based encoders, however, the number of tokens scales quadratically with resolution, making high-resolution processing prohibitive for tasks such as image search (\Cref{sec:retrieval}) and video query localization (\Cref{sec:videos}), which require caching representations for large collections of images or frames. Although \tren{} also relies on a patch-based backbone and therefore incurs higher computation when processing individual high-resolution images, it stores and propagates only aggregated region tokens for downstream tasks. This design keeps the number of cached tokens nearly constant regardless of the input resolution (see \Cref{fig:token-resolution}), allowing \tren{} to benefit from higher resolution with minimal additional storage overhead.

\vspace{0.6em}
\boldheader{Generalization to unseen categories}
In \Cref{sec:segmentation}, we evaluate \tren{} on ADE20K, which is not used during training. However, \tren{} is trained on a mixture of segmentation datasets containing over 4,600 category labels (\Cref{sec:training}), some of which overlap with the 150 ADE20K classes. To isolate true generalization, we identify five ADE20K categories that are entirely unseen during training (including synonyms): {\em conveyor belt}, {\em hovel}, {\em swivel chair}, {\em television receiver}, and {\em arcade machine}. We further identify 13 categories that appear in the training corpus only under different synonym forms. Evaluating performance on these subsets allows us to assess whether \tren{} preserves the open-vocabulary capabilities of the underlying DINOv3 text encoder. As shown in \Cref{fig:generalization}, \tren{} consistently outperforms DINOv3 dino.txt on these selected categories, mirroring its gains across the full 150-class benchmark.

% \vspace{0.6em}
% \boldheader{Cost analysis}

%% file: tables/visual-haystacks.tex
\begin{table}[t]
    \captionsetup{skip=5pt}
    \centering
    \caption{{\bf Visual Haystacks' single-needle challenge.} \tren{} outperforms open-source LMMs, RAG-based methods, and other vision-language encoders. ``E'' indicates context overflow, execution failure, or API error.}
    \vspace{-0.8em}
    \setlength{\tabcolsep}{3pt}
    \renewcommand{\arraystretch}{1.}
    \resizebox{0.98\linewidth}{!}{%
    \begin{tabular}{lcccccccccc}
        \toprule
        \textbf{Model} & \textbf{D=1} & \textbf{D=2} & \textbf{D=3} & \textbf{D=5} & \textbf{D=10} & \textbf{D=20} & \textbf{D=50} & \textbf{D=100} & \textbf{D=500} & \textbf{D=1K} \\
        \midrule
        Detector Oracle & 90.2 & 89.6 & 88.8 & 88.3 & 86.9 & 85.4 & 81.7 & 77.5 & 74.8 & 73.9 \\
        \midrule
        \multicolumn{11}{c}{{\em Proprietary LMMs}} \\
        \midrule
        Gemini-3 Pro & 88.9 & 89.2 & 87.3 & 87.2 & 85.7 & 83.5 & 74.3 & 74.1 & 71.0 & 67.9 \\
        Gemini-1.5 Pro & 88.4 & 82.0 & 78.3 & 76.0 & 71.9 & 68.6 & 62.8 & 57.4 & E & E \\
        GPT-4o & 82.5 & 79.9 & 77.5 & 73.3 & 68.2 & 65.4 & 59.7 & 55.3 & E & E \\
        \midrule
        \multicolumn{11}{c}{{\em Open-source LMMs}} \\
        \midrule
        LongVILA & 63.8 & 59.0 & 57.7 & 56.7 & 55.6 & 52.0 & 52.0 & 52.0 & E & E \\
        Qwen-2-VL & 80.9 & 76.6 & 73.6 & 67.9 & 62.6 & 59.1 & 52.6 & E & E & E \\
        Phi-3 & 80.5 & 69.1 & 67.3 & 62.0 & 54.8 & 52.6 & 50.8 & E & E & E \\
        InternVL-2 & 88.1 & 80.5 & 72.3 & 63.9 & 58.8 & 55.2 & E & E & E & E \\
        mPLUG-OWL3 & 84.4 & 66.0 & 62.1 & 57.0 & 53.2 & 51.5 & E & E & E & E \\
        \midrule
        \multicolumn{11}{c}{{\em Retrieval-Augmented Methods}} \\
        \midrule
        LLaVA-v1.5 & 85.8 & 77.1 & 75.8 & 68.6 & 63.6 & 60.4 & 55.3 & 57.5 & 55.4 & 52.9 \\
        MIRAGE & 83.2 & 77.8 & 76.6 & 72.8 & 70.5 & 66.0 & 63.6 & 62.0 & 58.7 & 55.7 \\
        \midrule
        \multicolumn{11}{c}{{\em Vision-Language Encoders}} \\
        \midrule
        SigLIP-2 & 72.0 & 69.2 & 68.1 & 65.3 & 64.1 & 60.3 & 58.7 & 58.3 & 56.6 & 54.9 \\
        REN & 81.2 & 78.6 & 77.4 & 76.0 & 74.0 & 72.1 & 68.3 & 65.5 & 62.3 & 59.2 \\
        DINOv3 dino.txt & 72.7 & 71.3 & 69.2 & 68.2 & 66.1 & 63.2 & 60.9 & 60.2 & 56.4 & 52.1 \\
        \rowcolor{blue}
        \highlight{\tren{}} & \highlight{88.5} & \highlight{86.4} & \highlight{85.3} & \highlight{83.9} & \highlight{82.6} & \highlight{79.6} & \highlight{75.2} & \highlight{74.0} & \highlight{68.2} & \highlight{65.2} \\
        \rowcolor{blue}
        \small$\Delta_\text{vs.\ DINOv3 dino.txt}$ & \small\gain{15.8} & \small\gain{15.1} & \small\gain{16.1} & \small\gain{15.7} & \small\gain{16.5} & \small\gain{16.4} & \small\gain{14.3} & \small\gain{13.8} & \small\gain{11.8} & \small\gain{13.1} \\
        \bottomrule
    \end{tabular}}
    \label{tab:haystacks}
    \vspace{-1.5em}
\end{table}

%% file: tables/retrieval-and-segmentation.tex
\begin{figure}[t]
    \centering
    \begin{minipage}[t]{0.48\textwidth}
        \centering
        \includegraphics[width=\linewidth]{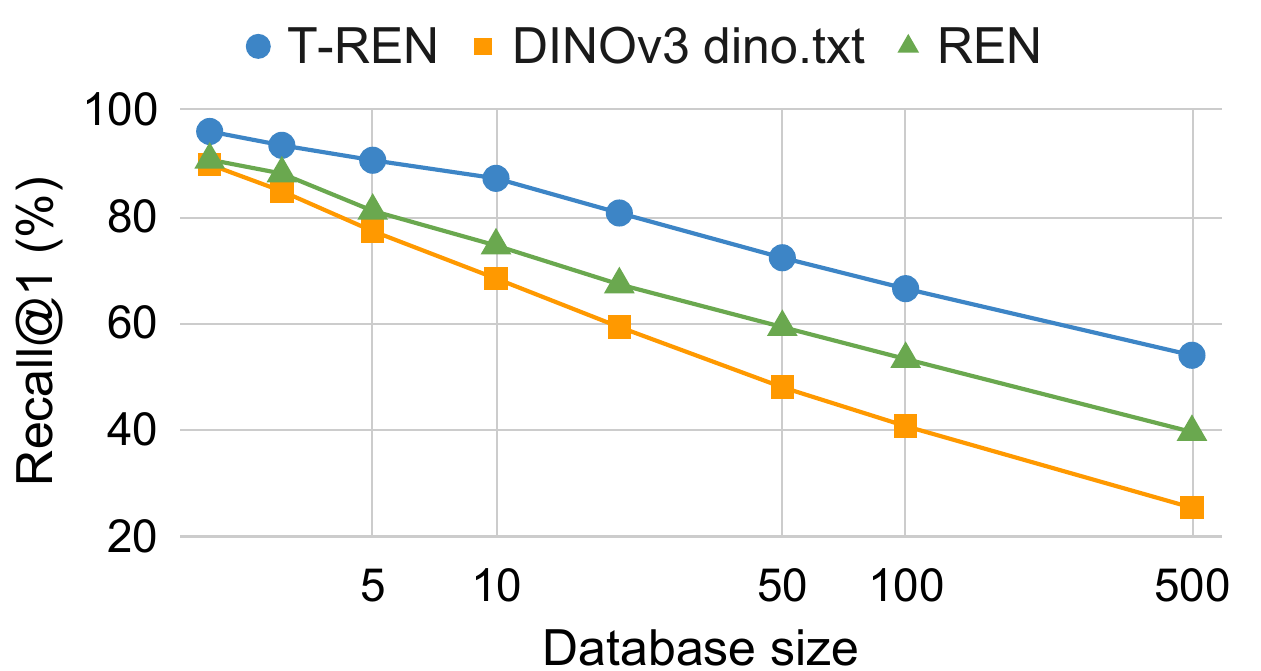}
        \vspace{-1.8em}
        \captionof{figure}{{\bf Zero-shot retrieval.} \tren{} improves R@1 by 18.4\% over DINOv3 dino.txt with $24\times$ fewer tokens, and by 10.8\% over REN, highlighting the advantage of multi-region token prediction.}
        \label{fig:text-based-retrieval}
        \vspace{0.8em}
        \includegraphics[width=0.9\linewidth]{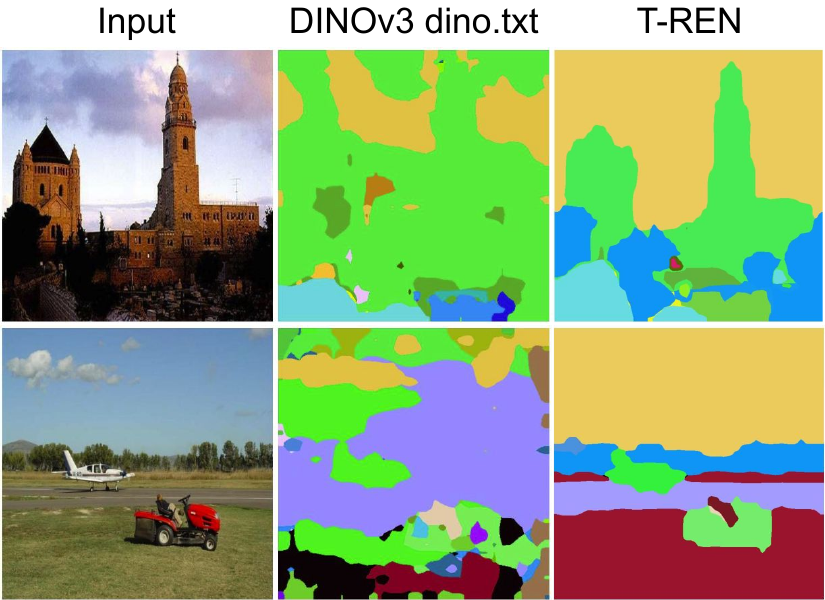}
        \vspace{-0.8em}
        \captionof{figure}{{\bf Qualitative OVSS results.} \tren{}'s segmentations better adhere to object boundaries than DINOv3 dino.txt's.}
        \label{fig:segmentation-examples}
    \end{minipage}
    \hfill
    \begin{minipage}[t]{0.49\textwidth}
        \vspace{-10.5em}
        \centering
        \captionof{table}{{\bf Open-vocabulary semantic segmentation.} \tren{} outperforms both patch-based encoders and SAM-guided methods. %Following~\cite{}, 
        \tren{} and patch-based approaches use a ViT-L backbone with 576 tokens per image. We additionally report higher-resolution results (\tren{}+) for a fairer comparison with SAM-guided methods, which use a ViT-H backbone and input resolutions of 672p and 1344p for ADE20k and Cityscapes, respectively.}
        \setlength{\tabcolsep}{3pt}
        \renewcommand{\arraystretch}{1.}
        \resizebox{0.97\linewidth}{!}{%
        \begin{tabular}{lcc}
            \toprule
            \textbf{Model} & \textbf{ADE20K} & \textbf{Cityscapes} \\
            \midrule
            \multicolumn{3}{c}{{\em SAM-guided approaches}} \\
            \midrule
            Trident & 25.6 & 46.9 \\
            RADSeg+ & 29.9 & 45.8 \\
            TextRegion & 27.3 & 47.4 \\
            \midrule
            \multicolumn{3}{c}{{\em Patch-based vision-language encoders}} \\
            \midrule
            CLIP & 6.0 & 11.5 \\
            EVA-02-CLIP & 10.9 & 14.1 \\
            SigLIP-2 & 10.8 & 16.3 \\
            PE & 17.6 & 21.4 \\
            DINOv2 dino.txt & 19.2 & 27.4 \\
            DINOv3 dino.txt & 24.7 & 36.9 \\
            \midrule
            \rowcolor{blue}
            \highlight{\tren{}} & \textbf{30.6} & \textbf{52.7} \\
            \rowcolor{blue}
            \highlight{\tren{}+} & \textbf{32.0} & \textbf{58.7} \\
            \bottomrule
        \end{tabular}}
        \label{tab:semantic-segmentation}
    \end{minipage}
    \vspace{-1.5em}
\end{figure}

%% file: tables/video-tasks.tex
\begin{table}[t]
    \centering
    \caption{{\bf Video tasks.} \tren{} improves performance on both video query localization and video scene parsing while significantly compressing the representation. For query localization, following \cite{Ego4D}, a prediction is considered correct localization if the temporal-IoU between the predicted and target temporal window exceeds 0.25.}
    \vspace{-0.8em}
    \setlength{\tabcolsep}{4pt}
    \renewcommand{\arraystretch}{1.1}
    \label{tab:video-tasks}
    \resizebox{0.92\linewidth}{!}{%
    \begin{tabular}{lccccc}
        \toprule
        \multirow{2}{*}{\textbf{Model}} & \multicolumn{3}{c}{\textbf{Query Localization (Ego4D)}} & \multicolumn{2}{c}{\textbf{Scene Parsing (VSPW)}} \\
        \cmidrule(lr){2-4} \cmidrule(lr){5-6}
        & \textbf{Recall@1} & \textbf{tAP} & \textbf{Compression ($\uparrow$)} & \textbf{mIoU} & \textbf{Compression ($\uparrow$)} \\
        \midrule
        DINOv3 dino.txt & 36.8 & 14.4 & 1$\times$ & 20.7 & 1$\times$ \\
        REN & 39.0 & 19.9 & 26.8$\times$ & 18.5 & 22.9$\times$ \\
        \rowcolor{blue}
        \highlight{\tren{}} & \highlight{52.4} & \highlight{26.4} & \highlight{187.5$\times$} & \highlight{38.3} & \highlight{254.5$\times$} \\
        \bottomrule
    \end{tabular}}
    \vspace{-1.5em}
\end{table}

%% file: tables/ablation.tex
\begin{table}[t]
    \centering
    \caption{{\bf Ablation on region pooling and text alignment.} Removing either region pooling or text alignment leads to degraded performance, demonstrating that both components are essential for \tren{}. ``VH'' denotes Visual Haystacks.}
    \label{tab:ablation}
    \vspace{-0.8em}
    \setlength{\tabcolsep}{4pt}
    \renewcommand{\arraystretch}{1.0}
    \resizebox{0.98\linewidth}{!}{%
    \begin{tabular}{cc | cccc}
        \toprule
        \textbf{Train Region} & \textbf{Train Text} & \textbf{ADE20K} & \textbf{Cityscapes} & \textbf{VH Retrieval} & \textbf{VH Reasoning} \\
        \textbf{Pooling} & \textbf{Alignment} & \textbf{(mIoU)} & \textbf{(mIoU)} & \textbf{(D=10)} & \textbf{(D=10)} \\
        \midrule
        &  & 24.7 & 36.9 & 68.4 & 66.1 \\
        & \cmark & 25.4 & 44.7 & 76.1 & 72.7 \\
        \cmark &  & 19.5 & 21.1 & 65.5 & 68.8 \\
        \rowcolor{blue}
        \cmark & \cmark & \textbf{30.6} & \textbf{52.7} & \textbf{87.2} & \textbf{82.6} \\
        \bottomrule
    \end{tabular}}
    \vspace{-1.5em}
\end{table}

%% file: tables/merging-effect.tex
\begin{figure}[t]
    \centering
    \begin{minipage}[t]{0.51\textwidth}
        \centering
        \captionof{table}{{\bf Ablation on token merging strategies.} In-frame merging alone achieves 29.2$\times$ compression with no mIoU loss; adding temporal merging reaches 254.5$\times$ compression at a cost of only 0.3 mIoU. Compression ratios are relative to a patch-based encoder baseline.}
        \vspace{0.5em}
        \setlength{\tabcolsep}{6pt}
        \renewcommand{\arraystretch}{1.1}
        \label{tab:merging-effect}
        \resizebox{\linewidth}{!}{%
        \begin{tabular}{cccc}
            \toprule
            \shortstack{\textbf{In-Frame} \\ \textbf{Merging}} 
            & \shortstack{\textbf{Temporal} \\ \textbf{Merging}}
            & \shortstack{\textbf{VSPW} \\ \textbf{mIoU}} 
            & \shortstack{\textbf{Comp.} \\ \textbf{($\uparrow$)}} \\
            \midrule
            &  & 38.6 & 1$\times$ \\
            \cmark &  & 38.6 & 29.2$\times$ \\
            \rowcolor{blue}
            \cmark & \cmark & \highlight{38.3} & \highlight{254.5$\times$} \\
            \bottomrule
        \end{tabular}}
    \end{minipage}
    \hfill
    \begin{minipage}[t]{0.46\textwidth}
        \vspace{1.2em}
        \centering
        \includegraphics[width=\linewidth]{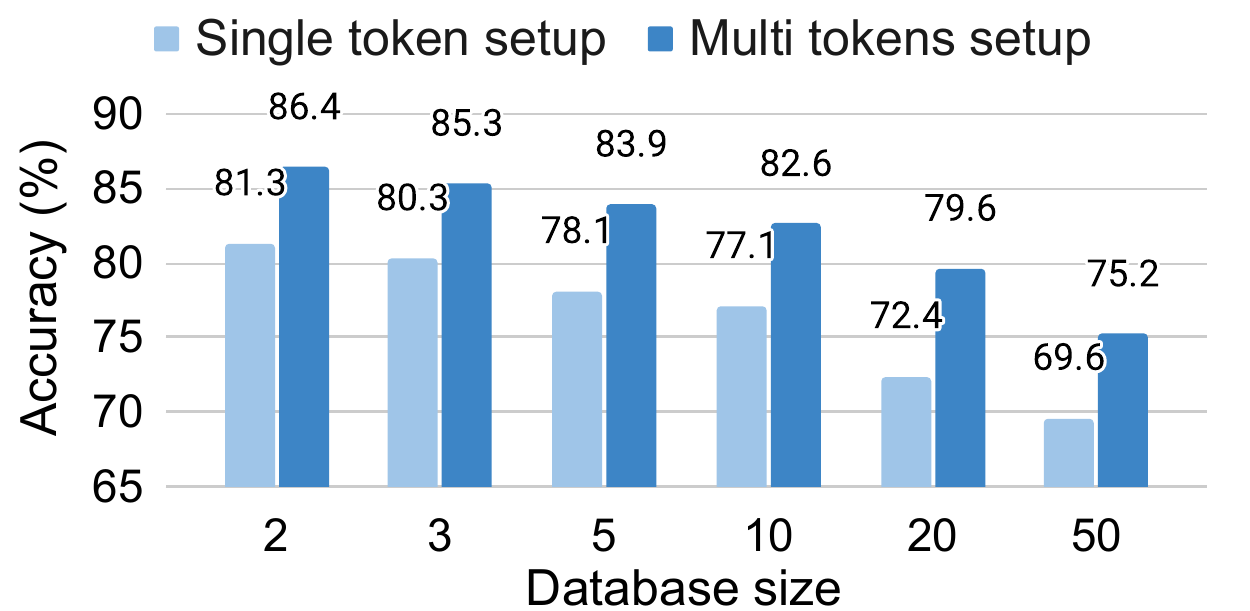}
        \vspace{-1.8em}
        \captionof{figure}{{\bf Single vs.\ multi-token setups.} Predicting multiple tokens per point consistently improves performance on Visual Haystacks, highlighting the benefit of capturing hierarchical visual structure.}
        \label{fig:multi-token}
    \end{minipage}
    \vspace{-1.5em}
\end{figure}

%% file: sections/5-conclusion.tex
\vspace{-0.5em}
\section{Conclusion}
\label{sec:conclusion}
\vspace{-0.5em}
We present \tren{}, a vision-language encoder that learns text-aligned region tokens by jointly pooling patch features into region tokens and aligning them with language. This design produces compact representations while enabling fine-grained cross-modal grounding. As a result, \tren{} supports both dense visual understanding and scalable representation for large visual collections. We demonstrate these advantages across diverse open-vocabulary settings, including image-level dense prediction, large-scale retrieval, and long-video parsing, supported by extensive ablations and analysis. Overall, \tren{} shows that text-aligned region tokens provide a principled and scalable foundation for open-vocabulary vision-language modeling.

\boldheader{Future direction}
While \tren{} leverages strong pretrained vision-language backbones, future work may explore training region-based vision-language models end-to-end to further strengthen this paradigm.

%% file: sections/6-supplementary.tex
% % Mention all 5+13 ADE20k categories for generalization experiment
% % Add FPS for streaming video processing in compute req

\clearpage
\appendix
\section{Supplementary Material}

\noindent{}This supplementary material is organized as follows: \Cref{sec:sensitivity-analysis} analyzes \tren{}'s sensitivity to key hyperparameters; \Cref{sec:compute-req} compares the computational requirements of \tren{} with baselines; and \Cref{sec:implementation} provides implementation and training details.

\subsection{Hyperparameter Sensitivity Analysis}
\label{sec:sensitivity-analysis}

\boldheader{Prompt grid size}
For all experiments in \Cref{sec:experiments}, we use a prompt grid matching the backbone patch tokens, i.e., $\frac{H}{16} \times \frac{W}{16}$ grid for images of resolution $H \times W$ with patch size 16. However, \tren{} can also be prompted with denser or sparser grids. We show the effect of varying the grid size on ADE20k and Visual Haystacks in \Cref{tab:prompt-grid-variation}, and find that \tren{} consistently outperforms DINOv3 dino.txt across all grid sizes, with performance remaining reasonably stable. Consequently, for compute-sensitive applications, we can use \tren{} with sparser grid to reduce the compute requirements (see \Cref{tab:compute-req}), while still maintaining superior performance and fewer tokens than DINOv3 dino.txt.

\input{tables/prompt-grid-variation}

\vspace{0.6em}
\boldheader{Visual token merging threshold}
We analyze the effect of the merging threshold $\tau_\text{mask}$ in \Cref{tab:iou-similarity-variation}. The threshold controls the degree of token merging in an image: lower values produce fewer tokens through aggressive merging, while higher values retain more regions and increase the token budget (\Cref{sec:approach}). As the threshold increases, both semantic segmentation (ADE20k~\cite{ADE20k}) and finding needle in a haystack reasoning (Visual Haystacks~\cite{VisualHaystacks}) improves steadily, reflecting the benefit of preserving finer spatial structure. Notably, \tren{} outperforms DINOv3 dino.txt with significantly fewer tokens; see \Cref{fig:count-vs-perf} that plots task performance vs.\ the average number of tokens needed to represent an image.

\input{tables/iou-similarity-variation}

\input{tables/track-merging-threshold}

\begin{figure}[t]
    \centering
    \includegraphics[width=\linewidth]{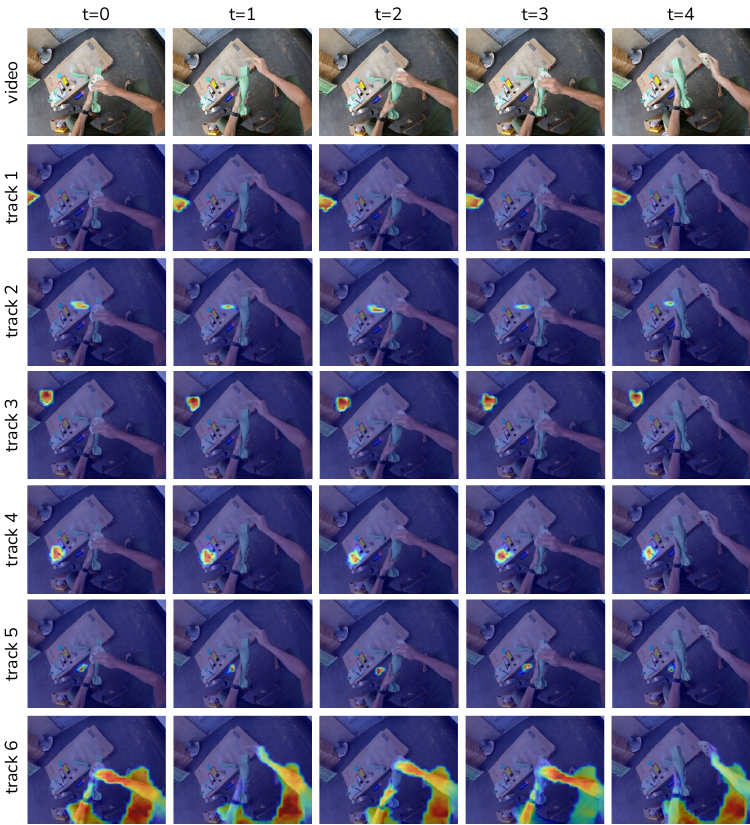}
    \vspace{-1.6em}
    \caption{{\bf Qualitative examples of region tracks.} The first row shows the video frames, and subsequent rows visualize the cross-attention masks of frame-level region tokens that are grouped into the same track. Our approach tracks both small objects amidst clutter (e.g., tracks 2, 3, 4, and 5) and larger objects (track 6).}
    \label{fig:track-tokens-1}
    \vspace{-1em}
\end{figure}

\begin{figure}[t]
    \centering
    \includegraphics[width=\linewidth]{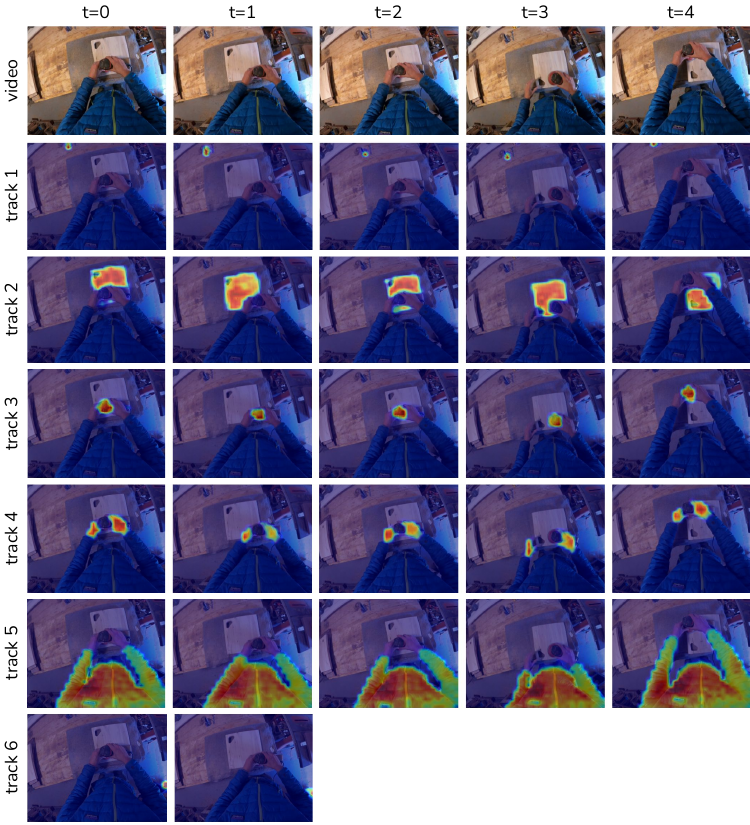}
    \vspace{-1.6em}
    \caption{{\bf Qualitative examples of region tracks.} \tren{} tracks extremely small objects (track 1), objects under occlusion (track 2), disjoint regions (track 4), and objects with brief and partial visibility (track 6).}
    \label{fig:track-tokens-2}
    \vspace{-1em}
\end{figure}

\begin{figure}[t]
    \centering
    \includegraphics[width=\linewidth]{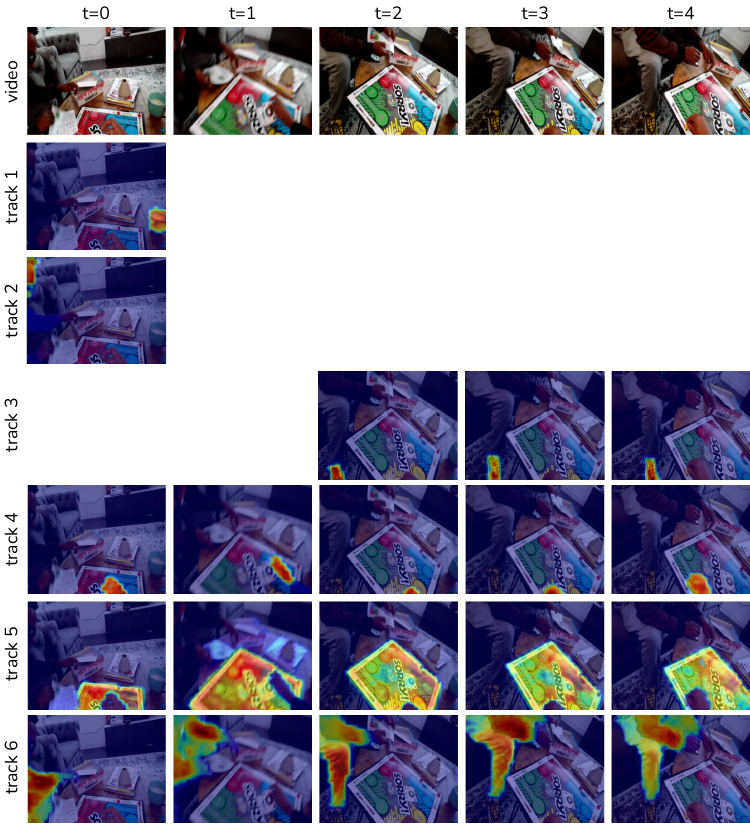}
    \vspace{-1.6em}
    \caption{{\bf Qualitative example of region tracks.} \tren{} is robust to partial visibility (e.g., track 4) and large viewpoint shifts (e.g., track 6, where the person's appearance changes significantly from $t=0$ to $t=4$). Track tokens span only the frames in which the object is visible: for instance, the tracks for the cup (track 1) and the person's face (track 2) appear only at $t=0$, while the shoe track (track 3) spans $t=2$ to $t=4$.}
    \label{fig:track-tokens-3}
    \vspace{-1em}
\end{figure}

\vspace{0.6em}
\boldheader{Temporal merging threshold}
We analyze the effect of the temporal merging threshold $\tau_\text{track}$ on video query localization in \Cref{tab:track-merging-threshold}. Performance remains stable for moderate values of $\tau_\text{track}$ (0.4--0.7). High values make track formation overly strict, preventing associations across frames when objects undergo large viewpoint changes or are only partially visible. Nevertheless, even with very high $\tau_\text{track}$ (or even without temporal token merging), \tren{} still outperforms DINOv3 dino.txt, highlighting the advantage of text-aligned region tokens over patch tokens. At very low values of $\tau_\text{track}$, track matching becomes more permissive and may introduce some spurious associations, though their impact is limited. Importantly, $\tau_\text{track}=0.1$ does not imply that any pair of tokens with similarity above 0.1 will be merged. Our implementation enforces greedy one-to-one matching between tokens in consecutive frames, ensuring that each token in frame $t+1$ is merged only with its most similar unmatched token in frame $t$ (see \Cref{sec:approach}). This constraint prevents widespread spurious merges and explains the robustness of \tren{} to $\tau_\text{track}$. Notably, even with ${\sim}300\times$ fewer tokens, \tren{} achieves ${\sim}13\%$ higher recall than DINOv3 dino.txt, suggesting that substantial temporal redundancy exists in real-world videos that \tren{} effectively exploits. Qualitative examples of tracked tokens are shown in \Cref{fig:track-tokens-1,fig:track-tokens-2,fig:track-tokens-3}.

\subsection{Compute Requirements}
\label{sec:compute-req}

We compare the computational cost of encoding a single $512 \times 512$ image using \tren{} against DINOv3-based dino.txt~\cite{DINOv3} and REN~\cite{REN}. Results are summarized in \Cref{tab:compute-req}, where we report three metrics: 

\vspace{0.6em}
\boldheader{Parameter count}
The vision encoder of \tren{} has a parameter count comparable to the patch-based DINOv3 dino.txt encoder (334.7M vs.\ 328.5M). In contrast, REN requires significantly more parameters (432.4M) because it relies on two separate backbones: one for region mask generation and another for extracting text-aligned features. Additionally, REN employs a less efficient cross-attention design for aggregating patch features. As a result, despite using multiple queries per point prompt, \tren{} requires fewer parameters than REN for pooling patch-level features into region tokens.

\vspace{0.6em}
\boldheader{Latency} 
We measure the wall-clock time required to process a single image and produce its visual representation. The additional region pooling and token merging operations in \tren{} introduce a modest increase in latency relative to the patch-based DINOv3 encoder. Importantly, the latency of \tren{} decreases as the number of point prompts is reduced. For example, with a $16 \times 16$ prompt grid, \tren{} requires only 1.7\,ms more than DINOv3 dino.txt to encode an image. Given the substantial compression in the resulting token representation and the improved downstream performance (see \Cref{tab:prompt-grid-variation}), this small latency overhead is practically negligible. Compared to REN, \tren{} remains consistently faster at encoding images. While REN also benefits from reduced prompt density, the latency reduction is less pronounced compared to \tren{}.

\vspace{0.6em}
\boldheader{FLOPs}
We report the number of floating-point operations (FLOPs) for a single forward pass of the visual encoder. With a $32 \times 32$ prompt grid, \tren{} incurs 23.9\% more FLOPs than DINOv3 dino.txt. However, as the prompt grid becomes sparser, this computational gap decreases substantially. For a $16 \times 16$ grid, \tren{} requires roughly the same FLOPs as DINOv3 dino.txt while achieving stronger performance across downstream tasks (see \Cref{tab:prompt-grid-variation}). These results demonstrate that \tren{} can produce compact and efficient region-based representations with minimal effect on computational cost.

\input{tables/compute-req}

\subsection{Implementation and Training Details}
\label{sec:implementation}

\begin{figure}[t]
    \centering
    \includegraphics[width=\linewidth]{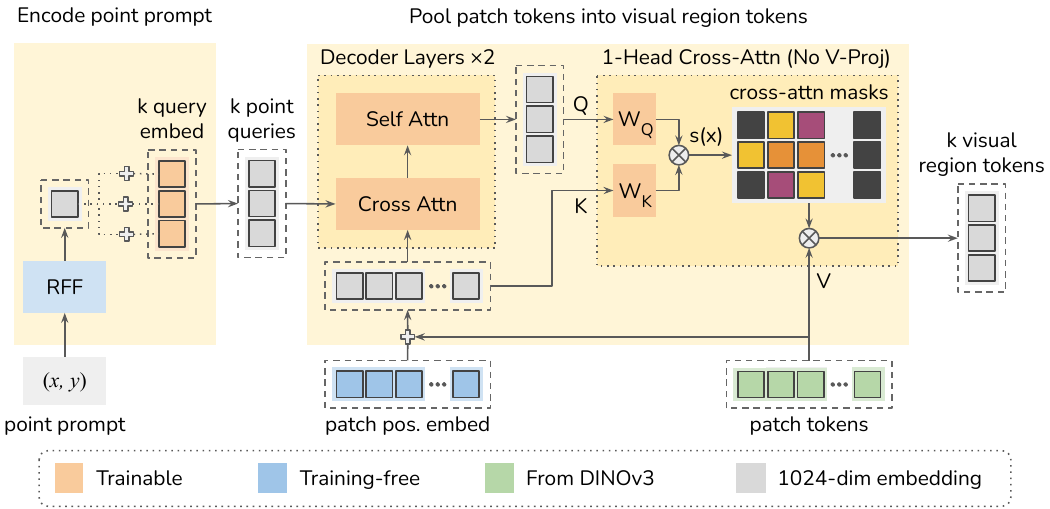}
    \vspace{-1.6em}
    \caption{{\bf Prompt encoding and region pooling.} Point prompt $(x, y)$ is mapped to a position embedding using Gaussian Random Fourier Features (RFF) and augmented with $k=3$ learnable query embeddings. $L=2$ decoder layers enrich the point queries with information of the patch tokens (via cross-attention) and other queries for the same point prompt (via self-attention). These contextually rich queries then attend to the patch tokens via a single-head cross-attention to produce visual region tokens (and cross-attention masks, as visualized in \Cref{fig:cross-attention-masks}). See \Cref{sec:implementation} for more details.}
    \label{fig:implementation}
    \vspace{-1em}
\end{figure}

\boldheader{Architecture}
\tren{}'s architecture is divided into the following components: 

\begin{enumerate}
    \item \textbf{Backbone.} We use DINOv3 ViT-L/16~\cite{DINOv3} to encode images into patch tokens. Specifically, an image of size $x \in \mathbb{R}^{H \times W \times 3}$ is converted into patch tokens $f \in \mathbb{R}^{N \times 1024}$, where $N = (H/16) \cdot (W/16)$. Text is encoded using the text encoder of DINOv3-based dino.txt~\cite{DINOtxt}.

    \item \textbf{Prompt encoder.} We first use Gaussian Random Fourier Features (RFF) to map $P$ point prompts into positional embeddings $e \in \mathbb{R}^{P \times 1024}$. Then, each of the $P$ positional embeddings is independently added to $k=3$ learnable query embeddings $z \in \mathbb{R}^{3 \times 1024}$ to produce point queries $q \in \mathbb{R}^{P \times 3 \times 1024}$. \Cref{fig:implementation} shows this for a single point prompt.
    
    \item \textbf{Decoder Layers.} As shown in \Cref{fig:implementation}, the decoder consists of a stack of $L=2$ Transformer layers, each composed of a standard cross-attention block followed by a self-attention block. For the cross-attention operation, the keys and values are obtained from the patch tokens augmented with positional encodings, denoted as $f^{\text{+pe}} \in \mathbb{R}^{N \times 1024}$. The queries for the first decoder layer are initialized using the point queries $q \in \mathbb{R}^{P \times 3 \times 1024}$ obtained from the prompt encoder. For subsequent layers, the queries are obtained by adding the positional embeddings of the corresponding point prompts ($e \in \mathbb{R}^{P \times 1024}$) to the output of the previous decoder layer. The output of the cross-attention block is then processed by a self-attention block to enable interaction among the query tokens, followed by a LayerNorm operation. Thus, the decoder layers output contextually enriched query tokens $q^\text{+} \in \mathbb{R}^{P \times 3 \times 1024}$ that incorporate information from image patch features via cross-attention and from other queries via self-attention. Both the cross-attention and self-attention modules use multi-head attention with 8 heads. 

    \item \textbf{Single-Head Cross-Attention.} Visual region tokens are generated via a cross-attention layer that uses the outputs of the decoder layers $q^\text{+} \in \mathbb{R}^{P \times 3 \times 1024}$ as queries, position augmented patch tokens $f^{\text{+pe}} \in \mathbb{R}^{N \times 1024}$ as keys, and original patch tokens $f \in \mathbb{R}^{N \times 1024}$ as values. It uses a single attention head and omits value projection as well as output projection. This produces visual region tokens $r^{(v)} \in \mathbb{R}^{P \times 3 \times 1024}$ and cross-attention masks $a \in \mathbb{R}^{P \times 3 \times N}$.

    \item \textbf{Merge.} The visual region tokens are merged if their pairwise cosine similarity exceeds $\tau_\text{token}=0.975$ or their cross-attention mask IoU exceeds $\tau_\text{mask}=0.8$, as described in Section~\ref{sec:method}. For video, track association uses $\tau_\text{track}{=}0.65$.

    \item \textbf{Text projector.} A two-layer MLP ($1024 \to 2048 \to 1024$, GELU, dropout $p = 0.1$) projects region tokens into the backbone's text embedding space.
\end{enumerate}

\vspace{0.6em}
\boldheader{Training data}
T-REN is trained on training splits of COCOStuff~\cite{COCOStuff}, OpenImagesV7~\cite{OpenImagesv7}, PhraseCut~\cite{PhraseCut}, Mapillary Vistas~\cite{Mapillary}, and SA-1B~\cite{SAM}. SA-1B does not contribute to $\mathcal{L}_\text{cont}^{(t)}$ (\cref{eq:txt-cont-loss}) as it does not have category labels; other datasets provide the supervision signal for $\mathcal{L}_\text{cont}^{(t)}$, but rarely have overlapping region masks needed for training multi-token prediction. For training, images are resized to $512 \times 512 \times 3$ via bicubic interpolation and masks are resized to $512\times512$ via nearest-neighbor interpolation. 128 point prompts are sampled from the locations covered by the ground-truth masks. Sampling probability is proportional to the squared number of overlapping ground-truth masks at each location. To ensure a robust region-text contrastive loss, we group synonyms and highly similar phrases and exclude them from the negative set of $\mathcal{L}_\text{cont}^{(t)}$. Specifically, text embeddings for all region categories in the training set are computed using the all-mpnet-base-v2 sentence transformer, and categories with cosine similarity greater than 0.725 are clustered together. For our training set, we obtain 2743 category clusters.

\vspace{0.6em}
\boldheader{Optimization}
We use AdamW with a learning rate of $0.001$ and a weight decay of $0.01$. We use a linear warmup over 1,500 steps followed by cosine decay to $0.5\times$. Training runs for 60,000 iterations ($<1$ epoch) with a batch size of 16, and we apply gradient clipping with a maximum norm of 5.0. The total loss is given by:
\begin{equation}
    \mathcal{L} = \mathcal{L}_\text{cont}^{(v)} + \mathcal{L}_\text{cont}^{(t)} + \mathcal{L}_\text{dist} + \mathcal{L}_\text{attn},
\end{equation}
corresponding to \cref{eq:vis-cont-loss,eq:txt-cont-loss,eq:txt-dist-loss,eq:attn-mask-loss} in the main paper. Both contrastive objectives use temperature $\tau=0.1$.


%% file: tables/prompt-grid-variation.tex
\begin{table}[!b]
    \centering
    \caption{{\bf Effect of prompt grid size on performance.} \tren{} consistently outperforms DINOv3 dino.txt, irrespective of the prompt grid size. For these experiments, we only vary the prompt grid size, while all other hyperparameters are kept constant. To keep our analysis consistent with evaluations in \Cref{sec:experiments}, we use an image resolution of 384p and no token merging for ADE20k, and an image resolution of 512p with token merging for Visual Haystacks (VH). The token compression is reported for the VH evaluation set.}
    \vspace{-0.8em}
    \setlength{\tabcolsep}{4pt}
    \renewcommand{\arraystretch}{1.}
    \label{tab:prompt-grid-variation}
    \resizebox{0.84\linewidth}{!}{%
    \begin{tabular}{lcccc}
        \toprule
        \multirow{2}{*}{\textbf{Model}} & \textbf{Prompt} & \textbf{ADE20k} & \textbf{VH (D=10)} & \textbf{Spatial Token} \\
        & \textbf{Grid Size} & \textbf{mIoU} & \textbf{Accuracy} & \textbf{Compression} \\
        \midrule
        DINOv3 dino.txt & n/a & 24.7 & 66.1 & 1$\times$ \\
        \midrule
        \multirow{6}{*}{\tren{}} & $16 \times 16$ & 29.2 & 81.4 & \textbf{29.1$\times$} \\
        & $20 \times 20$ & 29.9 & \textbf{83.0} & 26.6$\times$ \\
        & $24 \times 24$ & \textbf{30.6} & 81.6 & 25.7$\times$ \\
        & $32 \times 32$ & 30.4 & 82.6 & 24.6$\times$ \\
        & $48 \times 48$ & 30.4 & 78.2 & 30.4$\times$ \\
        & $64 \times 64$ & 30.2 & 76.1 & 35.2$\times$ \\
        \bottomrule
    \end{tabular}}
\end{table}

%% file: tables/iou-similarity-variation.tex
\begin{figure}[!th]
    \centering
    \begin{minipage}{\linewidth}
        \centering
        \captionof{table}{{\bf Effect of $\mathbf{\tau_\text{mask}}$ on performance.} 
        On both ADE20k and Visual Haystacks (VH), performance remains stable for $0.5 \leq \tau_\text{mask} \leq 0.9$, with only a 0.8 mIoU variation on ADE20k and 1\% accuracy variation on VH. The reported token count (number of tokens needed to represent one image) is averaged over the evaluation dataset. ``NTM'' denotes No Token Merging. The blue cells highlight the threshold value at which \tren{} starts performing comparably to DINOv3 dino.txt.}
        \vspace{0.4em}
        \setlength{\tabcolsep}{4pt}
        \renewcommand{\arraystretch}{1.}
        \resizebox{0.89\linewidth}{!}{%
        \begin{tabular}{lccccc}
            \toprule
            \multirow{2}{*}{\textbf{Model}} & \multirow{2}{*}{$\mathbf{\tau_{mask}}$} & \multicolumn{2}{c}{\textbf{ADE20k}} & \multicolumn{2}{c}{\textbf{VH (D=10)}} \\
            \cmidrule(lr){3-4} \cmidrule(lr){5-6}
            & & \textbf{mIoU} & \textbf{Token Count} & \textbf{Accuracy} & \textbf{Token Count} \\
            \midrule
            DINOv3 dino.txt & n/a & 24.7 & 576 & 66.1 & 1024 \\
            \midrule
            \multirow{11}{*}{\tren{}} & 0.0 & 8.1 & 1 & 55.7 & 1 \\
            & 0.1 & 17.5 & 9.8 & \cellcolor{blue}65.7 & \cellcolor{blue}7.7 \\
            & 0.2 & 21.9 & 15.5 & 71.1 & 14.1 \\
            & 0.3 & \cellcolor{blue}24.4 & \cellcolor{blue}20.4 & 76.2 & 20.0 \\
            & 0.4 & 26.4 & 26.8 & 80.0 & 27.5 \\
            & 0.5 & 27.4 & 32.8 & 81.5 & 34.8 \\
            & 0.6 & 27.6 & 34.4 & 81.9 & 37.1 \\
            & 0.7 & 27.8 & 36.3 & 82.0 & 39.9 \\
            & 0.8 & 28.1 & 37.4 & 82.6 & 41.6 \\
            & 0.9 & 28.2 & 37.8 & 82.5 & 42.3 \\
            & NTM & 30.6 & 576 & 83.5 & 1024 \\
            \bottomrule
        \end{tabular}}
        \label{tab:iou-similarity-variation}
    \end{minipage}

    \vspace{1em}

    \begin{minipage}{\linewidth}
        \centering
        \begin{subfigure}{0.49\linewidth}
            \includegraphics[width=\linewidth]{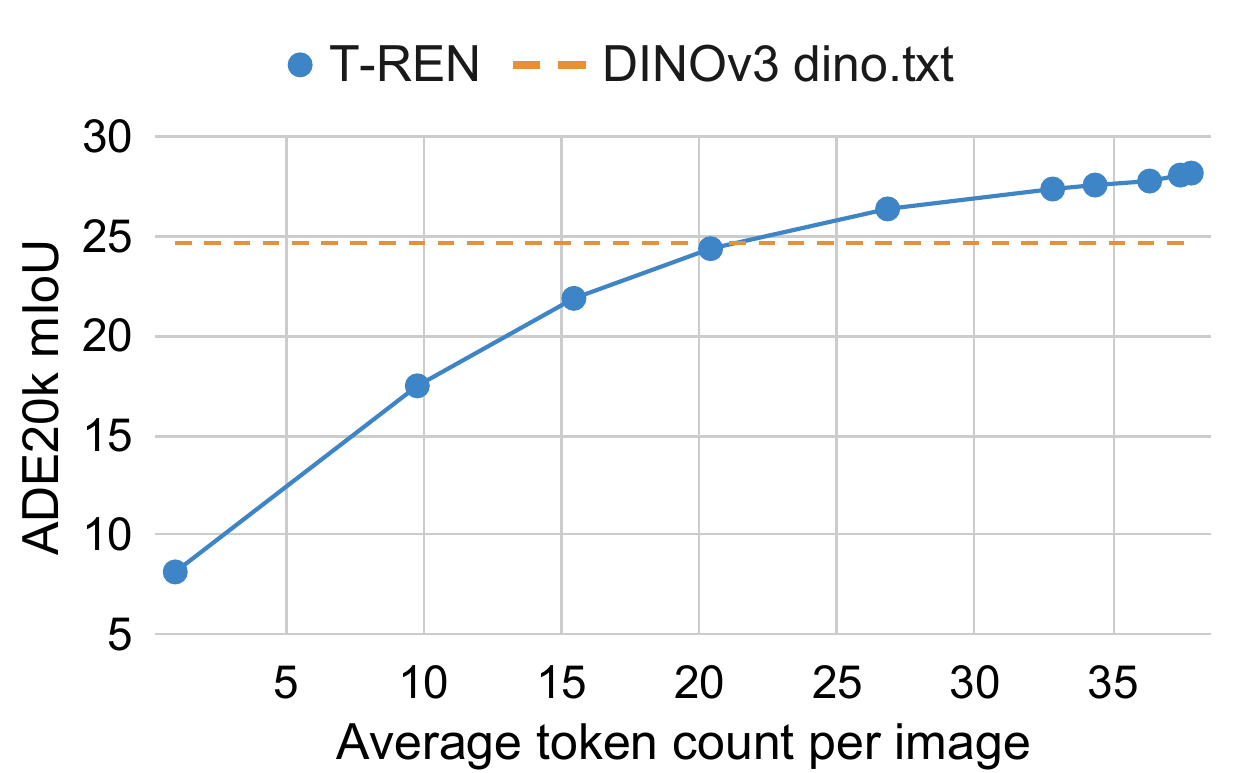}
            \label{fig:ade-vs-token-count}
        \end{subfigure}
        \hfill
        \begin{subfigure}{0.49\linewidth}
            \includegraphics[width=\linewidth]{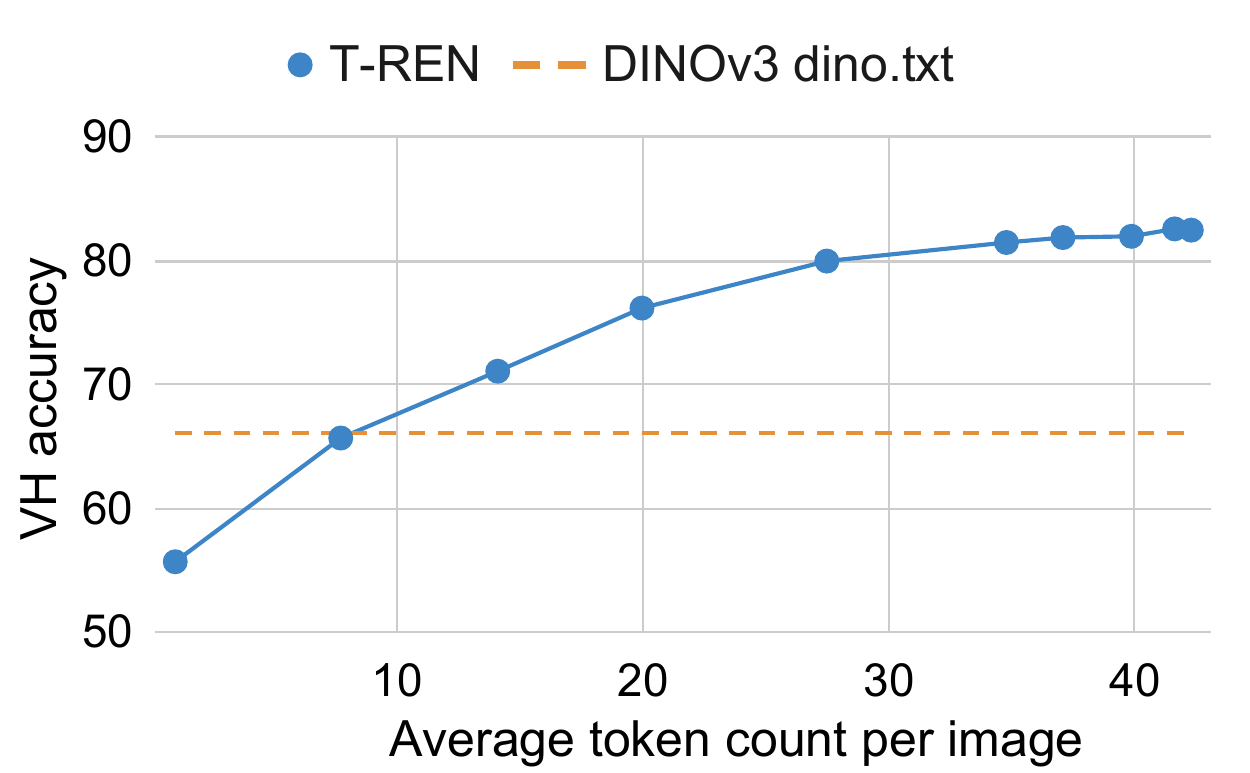}
            \label{fig:vh-vs-token-count}
        \end{subfigure}
        \vspace{-1.2em}
        \captionof{figure}{{\bf Performance vs.\ token count.} On ADE20k (left), \tren{} matches the performance of DINOv3 dino.txt using only 20 tokens per image (vs.\ 576 for DINOv3 dino.txt) and surpasses it with larger token budgets. On Visual Haystacks (right), \tren{} achieves comparable performance to DINOv3 dino.txt using just 8 tokens per image (vs.\ 1024 for DINOv3 dino.txt) and significantly outperforms it beyond that.}
        \label{fig:count-vs-perf}
    \end{minipage}
    \vspace{-1.5em}
\end{figure}

%% file: tables/track-merging-threshold.tex
% \begin{table}[t]
%     \centering
%     \caption{{\bf Effect of $\mathbf{\tau_\text{track}}$ on performance.} On VSPW, \tren{} maintains highly stable performance across a wide range of $\tau_\text{track}$ values, indicating low sensitivity to this hyperparameter. ``NTTM'' denotes No Temporal Token Merging; in this setting we disable temporal merging but still apply in-frame token merging with $\tau_\text{mask}=0.8$.}
%     \vspace{-0.8em}
%     \setlength{\tabcolsep}{4pt}
%     \renewcommand{\arraystretch}{1.}
%     \label{tab:track-merging-threshold}
%     \resizebox{0.7\linewidth}{!}{%
%     \begin{tabular}{lccc}
%         \toprule
%         \multirow{2}{*}{\textbf{Model}} & \multirow{2}{*}{$\mathbf{\tau_{track}}$} & \textbf{VSPW} & \textbf{Spatiotemporal} \\
%         & & \textbf{mIoU} & \textbf{Token Compression} \\
%         \midrule
%         DINOv3 dino.txt & n/a & 20.5 & 1$\times$ \\
%         \midrule
%         \multirow{10}{*}{\tren{}}
%         & 0.01 & \textcolor{red}{Add} & \textcolor{red}{Add} \\
%         & 0.1 & 40.1 & 125.7$\times$ \\
%         & 0.2 & 40.1 & 123.6$\times$ \\
%         & 0.3 & 40.4 & 115.5$\times$ \\
%         & 0.4 & 40.6 & 105.6$\times$ \\
%         & 0.5 & 40.8 & 94.9$\times$ \\
%         & 0.6 & 40.7 & 84.4$\times$ \\
%         & 0.7 & 40.9 & 74.2$\times$ \\
%         & 0.8 & 41.0 & 62.7$\times$ \\
%         & 0.9 & 41.0 & 47.7$\times$ \\
%         & NTTM & 40.9 & 25.2$\times$ \\
%         \bottomrule
%     \end{tabular}}
% \end{table}

\begin{table}[t]
    \centering
    \caption{{\bf Effect of $\mathbf{\tau_\text{track}}$ on performance.} On Ego4D video query localization, \tren{} maintains stable performance for $0.4 \leq \tau_\text{track} \leq 0.7$. ``NTTM'' denotes No Temporal Token Merging; in this setting we disable temporal merging but still apply in-frame token merging with $\tau_\text{mask}=0.8$. This analysis is conducted on a ${\sim}10\%$ subset of the Ego4D validation set (330 queries) for efficiency, as it involves sweeping across 10 values of $\tau_\text{track}$; the trends are nonetheless clear. Full validation set results are reported in \Cref{tab:video-tasks}.}
    \vspace{-0.8em}
    \setlength{\tabcolsep}{4pt}
    \renewcommand{\arraystretch}{1.}
    \label{tab:track-merging-threshold}
    \resizebox{0.95\linewidth}{!}{%
    \begin{tabular}{lcccc}
        \toprule
        % \multirow{2}{*}{\textbf{Model}} & \multirow{2}{*}{$\mathbf{\tau_{track}}$} & \multicolumn{2}{c}{\textbf{Localization Recall}} & \textbf{Spatiotemporal} \\
        % \cmidrule{3-4}
        % & & \textbf{tIoU=0.25} & \textbf{tIoU=0.05} & \textbf{Token Compression} \\
        \multirow{2}{*}{\textbf{Model}} & \multirow{2}{*}{$\mathbf{\tau_{track}}$} & \multicolumn{2}{c}{\textbf{Localization Recall}} & \multirow{2}{*}{\makecell{\textbf{Spatiotemporal}\\\textbf{Token Compression}}} \\
        \cmidrule{3-4}
        & & \textbf{tIoU=0.25} & \textbf{tIoU=0.05} & \\
        \midrule
        DINOv3 dino.txt & n/a & 35.8 & 52.1 & 1$\times$ \\
        \midrule
        \multirow{10}{*}{\tren{}}
        & 0.1 & 49.2 & 64.4 & 297.0$\times$ \\
        & 0.2 & 52.1 & 67.6 & 290.6$\times$ \\
        & 0.3 & 51.9 & 66.5 & 272.0$\times$ \\
        & 0.4 & 56.7 & 68.2 & 241.1$\times$ \\
        & 0.5 & 57.7 & 66.2 & 202.2$\times$ \\
        & 0.6 & 54.8 & 67.0 & 161.9$\times$ \\
        & 0.7 & 55.9 & 66.5 & 124.0$\times$ \\
        & 0.8 & 50.5 & 62.3 & 88.2$\times$ \\
        & 0.9 & 45.2 & 59.6 & 52.9$\times$ \\
        & NTTM & 39.4 & 55.6 & 22.7$\times$ \\
        \bottomrule
    \end{tabular}}
\end{table}

%% file: tables/compute-req.tex
\begin{table}[t]
    \centering
    \caption{{\bf Compute requirements for encoding visual input.} For \tren{} and REN, computational usage can be controlled by adjusting the prompt grid size, with smaller grids reducing the computational cost. All measurements are obtained on a single NVIDIA A40 GPU.}
    \vspace{-0.8em}
    \setlength{\tabcolsep}{6pt}
    \renewcommand{\arraystretch}{1.}
    \label{tab:compute-req}
    \resizebox{0.82\linewidth}{!}{%
    \begin{tabular}{lcccc}
        \toprule
        \multirow{2}{*}{\textbf{Model}} & \textbf{Prompt} & \textbf{Params} & \textbf{Latency} & \textbf{FLOPs} \\
         & \textbf{Grid Size} & \textbf{(M)} & \textbf{(ms)} & \textbf{(GFLOPs)} \\
        \midrule
        DINOv3 dino.txt & n/a & 328.5 & 67.7$\pm$0.2 & 787.93 \\
        \midrule
        REN & 32$\times$32 & 432.4 & 92.2$\pm$0.2 & 817.39 \\
        \rowcolor{blue}
        \tren{} & 32$\times$32 & 334.7 & 84.2$\pm$0.2 & 976.38 \\
        \midrule
        REN & 24$\times$24 & 432.4 & 87.3$\pm$0.2 & 785.44 \\
        \rowcolor{blue}
        \tren{} & 24$\times$24 & 334.7 & 74.8$\pm$0.1 & 862.54 \\
        \midrule
        REN & 16$\times$16 & 432.4 & 86.2$\pm$0.1 & 762.63 \\
        \rowcolor{blue}
        \tren{} & 16$\times$16 & 334.7 & 69.4$\pm$0.2 & 790.30 \\
        \bottomrule
    \end{tabular}}
\end{table}